%% file: main.tex
\documentclass[sigconf,nonacm]{acmart}

\usepackage{float}      
\usepackage{stfloats}   
\newsavebox{\teaserpanela}\newsavebox{\teaserpanelb}\newsavebox{\teaserpanelc}\newsavebox{\teaserrow} 
\newlength{\teaserht}
\usepackage{colortbl}   

\usepackage{xcolor}
\usepackage{xspace}
\usepackage{enumitem}
\usepackage{balance}    
\input{dfn}

\newcommand{\method}{\textsc{TANGCO}\xspace}
\newcommand{\methodpre}{\textsc{TANGCO}\textsuperscript{pre}\xspace}
\newcommand{\tlr}{\textsc{T-LR}\xspace}
\newcommand{\nognn}{No-GNN\xspace}

\AtBeginDocument{%
  }

\title{\method: Learning Topology-Aware Capacity Allocation for Overload-driven Cascading Failures}

\author{Orkun Irsoy}
\affiliation{%
  \institution{Carnegie Mellon University}
  \department{Electrical and Computer Engineering}
  \city{Pittsburgh}
  \state{Pennsylvania}
  \country{USA}}
\email{oirsoy@andrew.cmu.edu}

\author{Leman Akoglu}
\affiliation{%
  \institution{Carnegie Mellon University}
  \department{Heinz College}
  \city{Pittsburgh}
  \state{Pennsylvania}
  \country{USA}}
\email{lakoglu@andrew.cmu.edu}

\author{Osman Yagan}
\affiliation{%
  \institution{Carnegie Mellon University}
  \department{Electrical and Computer Engineering}
  \city{Pittsburgh}
  \state{Pennsylvania}
  \country{USA}}
\email{oyagan@andrew.cmu.edu}

\begin{abstract}
Many networked systems, from power grids to traffic networks and cloud clusters, carry loads across nodes with limited capacity.
A node whose load exceeds its capacity fails and sheds its load onto its neighbors, which can trigger a system-wide cascade.
We study how to allocate a fixed capacity budget across nodes to resist these cascades under local load redistribution.
The problem is difficult because no optimal allocation is known, and the fail-or-survive objective is non-differentiable and piecewise constant, so exact and gradient-based optimization methods do not directly apply.
We introduce \textsc{TANGCO} (Topology-Aware Neural Graph-Guided Capacity Optimization), which uses a graph neural network policy trained through the cascade simulator with policy-gradient learning and a heuristic anchor.
We evaluate \method on five synthetic graph families and five real networks spanning power, road, air, and Internet topologies.
The learned policy improves on the best of four hand-designed heuristics in all 450 synthetic instances and in 40 of 45 real-network conditions, with robustness gains ranging from 1.6\% to 246\%.
The learned policies transfer to unseen graphs within a family and partially across related topologies, and \methodpre, pre-trained on synthetic graphs, matches per-network training on unseen real networks.
Training scales near-linearly with graph size, and \methodpre allocates on a new network with no per-target training, matching the deployment cost of a hand-designed heuristic.
Free-vector variants without the GNN, trained by policy gradient or by CMA-ES, stay close to the heuristics, so the graph representation carries the gain beyond numerical search; the depth analysis shows most of it arises from one-hop information.
Finally, analysis of the learned allocations identifies when local risk is sufficient, leads to an improved closed-form heuristic, and reveals the regimes where a topology-aware learned policy remains necessary.
\end{abstract}



\begin{document}
\maketitle

\input{sections/introduction}
\input{sections/formulation}
\input{sections/method}
\input{sections/experiments}

\input{sections/scalability}
\input{sections/interpretability}

\input{sections/related_work}

\input{sections/conclusion}

\begin{acks}
This work was supported in part by the David Barakat and LaVerne Owen-Barakat Fellowship awarded to Orkun Irsoy by the Carnegie Mellon University College of Engineering.
This work was supported in part by the Air Force Office of Scientific Research (AFOSR) Grant \# FA9550-22-1-0233.
\end{acks}

\balance
\bibliographystyle{ACM-Reference-Format}
\bibliography{references}

\clearpage
\input{sections/appendix}

\end{document}

%% file: dfn.tex
\makeatletter
\newcommand\footnoteref[1]{\protected@xdef\@thefnmark{\ref{#1}}\@footnotemark}
\makeatother

\newcommand{\cbit}{\begin{compactitem}}
\newcommand{\ceit}{\end{compactitem}}
\newcommand{\cben}{\begin{compactenum}}
\newcommand{\ceen}{\end{compactenum}}

\newcommand{\beq}{\begin{equation}}
	\newcommand{\eeq}{\end{equation}}

\definecolor{darkgreen}{RGB}{41,166,41}

\newcommand{\bit}{\begin{itemize}}
	\newcommand{\eit}{\end{itemize}}
\newcommand{\ben}{\begin{enumerate}}
	\newcommand{\een}{\end{enumerate}}

\newcounter{x}

\definecolor{celadon}{rgb}{0.67, 0.88, 0.69}
\definecolor{carolinablue}{rgb}{0.6, 0.73, 0.89}

\definecolor{aliceblue}{rgb}{0.867, 0.917, 0.964}
\definecolor{aliceyellow}{rgb}{0.999, 0.945, 0.796}
\definecolor{alicegray}{rgb}{0.844, 0.867, 0.898}


%% file: sections/introduction.tex
\section{Introduction}
\label{sec:introduction}

Modern online services run on clusters of servers that share a request load.
When one server becomes overloaded, its share of the load shifts to the remaining servers; if their updated load with the addition exceeds their capacity, they fail in turn, triggering a cascade that can end in a system-wide collapse.
Such overload-driven cascades are a well-documented cause of large-scale outages at major cloud providers~\cite{bronson2021}.
The same pattern is prevalent across many networked systems, including power grids~\cite{andersson2005}, transportation and traffic networks~\cite{li_havlin2015,transportation_cascading_ex}, supply chains~\cite{supply_chain_cascade_ex,Global_trade_cascading_ex},
as well as financial systems where the failure of one institution imposes losses on its counterparties and can push them into failure~\cite{gai_kapadia2010}.
In each case a node carries a load and fails when that load exceeds its capacity, and a failed node passes its load to other functioning nodes.
The failure of a few components can therefore propagate through the network and bring down the system~\cite{buldyrev2010}.

Cascading failures are due to network effects, where entities are connected through dependencies in a graph.
The interaction between the graph topology and the capacity allocation of nodes therefore becomes critical in the cascade dynamics.
Flow-based models of cascading failure originate with Motter and Lai~\cite{MotterLai2002}, who take each node's load to be its betweenness centrality and set its capacity to a fixed multiple of that load, $c=(1+\alpha)l$.
A large literature builds on this model to relate network structure to robustness and to add mechanisms such as dynamical
recovery of failed nodes~\cite{majdandzic2014} and heterogeneous rules for how a
failed node's load spreads to its neighbors~\cite{hou2017heterogeneous}.
Almost all work in this line fixes the allocated capacity in advance and studies the robustness that results~\cite{MotterLai2002, wang_kim2007, li2008limited,chen2024loadcapacity}.
The capacity allocation is thus an input to the model rather than a quantity to be designed.

In this work, we treat capacity as a \emph{decision variable} and address the problem of allocating a limited capacity budget across nodes to maximize robustness against cascading failures, a question that has received far less attention.
The few works that study this allocation problem~\cite{zhang_optimizing,ozel2018uniform,irsoy_optimization_multiplex} solve it under a \emph{global} redistribution rule, where a failed node's load is shared equally among all surviving nodes, irrespective of graph topology.
That assumption yields a closed-form characterization of the final system size, and its allocations carry optimality guarantees while treating the network as fully connected and discarding topology.

Under \emph{local} redistribution, where loads pass only to immediate neighbors, cascade outcomes depend jointly on the topology and the initial load and capacity, and the global-redistribution guarantees no longer apply.
Existing allocations for \emph{local} redistribution are hand-designed heuristics rather than optimized methods.
Hence, even though prior work has studied robustness extensively by analyzing and modifying network structure, the design question of how to allocate capacities on a given network and load remains far less explored.
In contrast, we ask how to allocate a fixed capacity budget over a \textit{given} network under \textit{local} redistribution, where no optimal allocation is known.

We introduce \method (Topology-Aware Neural Graph-Guided Capacity Optimization), which employs a graph neural network (GNN) to learn capacity allocations directly from cascade dynamics.
Given a graph and its initial loads, \method outputs a capacity allocation that satisfies a fixed capacity budget.

Our paper introduces four main contributions:
\begin{itemize}
[leftmargin=10pt, itemsep=0.025in, topsep=0pt]

\item \textbf{Topology-aware capacity optimization:~}
We cast capacity allocation as a topology-aware policy-learning problem and propose \method to maximize robustness against cascading failures.
Hard overload thresholds make the cascade objective piecewise-constant; \method trains a GNN allocation policy from simulator AUC rewards with score-function updates, without differentiating through the discrete failures.

\item \textbf{Synthetic Pre-training and Transfer:~} 
We also introduce \methodpre, pretrained on diverse synthetic graph topologies and load distributions, and demonstrate strong transfer to unseen real-world networks (Figure~\ref{fig:teaser}).

\item \textbf{Effectiveness, Scalability, Speed:~} Across five graph families, three load distributions, three capacity budgets, and five real-world networks, \method consistently outperforms four competitive heuristics. Training scales linearly with graph size, while \methodpre needs no per-target training, yielding the best latency--performance trade-off (Figure~\ref{fig:teaser}). 

\item \textbf{Characterizing  Gains:~} 
We show that message passing contributes consistently beyond direct numerical optimization, develop a one-hop rule that recovers most gains in specific regimes, and find that \method's gains are largest on heterogeneous topologies at intermediate capacity budgets.

\end{itemize}

\savebox{\teaserpanela}{\resizebox{0.30\textwidth}{!}{%
\begin{tabular}{lccc}
\toprule
Network & Heur.* & \methodpre & \method \\
\midrule
\multicolumn{4}{l}{\textit{Uniform load}} \\
US grid     & 0.022 & 0.042 & 0.034 \\
Oregon AS   & 0.237 & 0.336 & 0.319 \\
Chicago     & 0.019 & 0.023 & 0.020 \\
OpenFlights & 0.076 & 0.261 & 0.262 \\
AS1221      & 0.263 & 0.239 & 0.286 \\
\midrule
\multicolumn{4}{l}{\textit{Pareto load}} \\
US grid     & 0.020 & 0.060 & 0.040 \\
Oregon AS   & 0.243 & 0.321 & 0.326 \\
Chicago     & 0.012 & 0.011 & 0.012 \\
OpenFlights & 0.082 & 0.213 & 0.240 \\
AS1221      & 0.260 & 0.250 & 0.282 \\
\bottomrule
\end{tabular}}}%
\setlength{\teaserht}{\dimexpr\ht\teaserpanela+\dp\teaserpanela\relax}%
\savebox{\teaserpanelb}{\includegraphics[height=\teaserht]{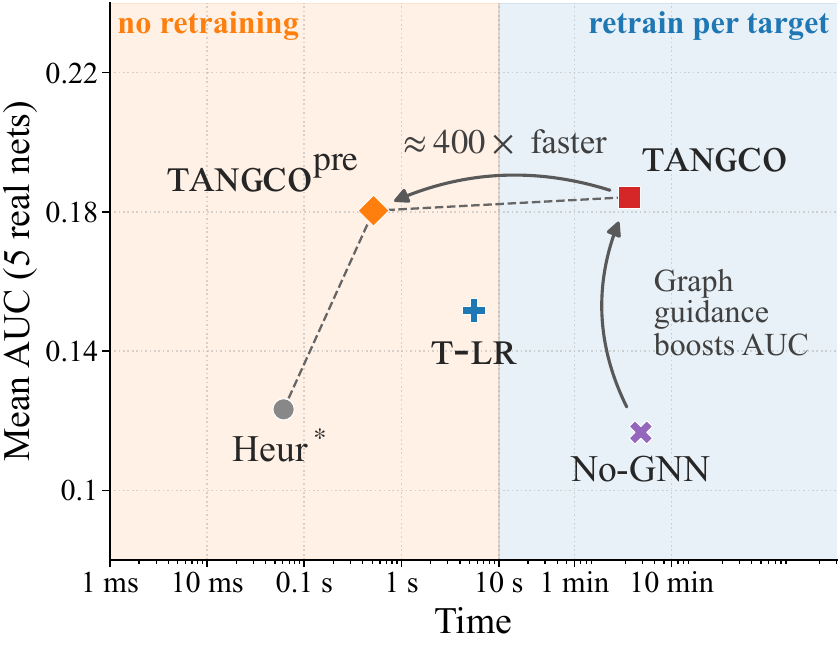}}%
\savebox{\teaserpanelc}{\includegraphics[height=\teaserht]{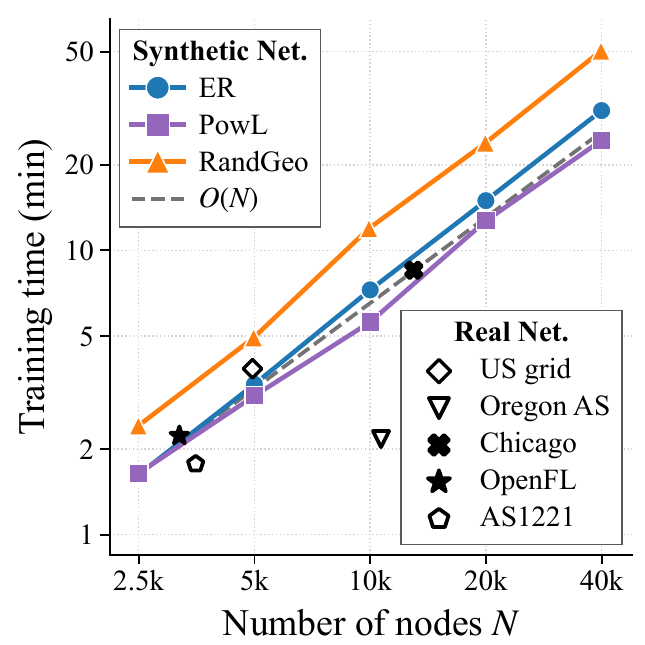}}%
\sbox{\teaserrow}{%
\begin{minipage}[b]{\wd\teaserpanela}
\centering
\usebox{\teaserpanela}
\\[3pt]
{\small (a) Robustness (AUC)}
\end{minipage}\hspace{0.6em}%
\begin{minipage}[b]{\wd\teaserpanelb}
\centering
\usebox{\teaserpanelb}
\\[3pt]
{\small (b) Robustness vs.\ deployment cost}
\end{minipage}\hspace{0.6em}%
\begin{minipage}[b]{\wd\teaserpanelc}
\centering
\usebox{\teaserpanelc}
\\[3pt]
{\small (c) Training time vs.\ network size}
\end{minipage}}%
\begin{figure*}[t]
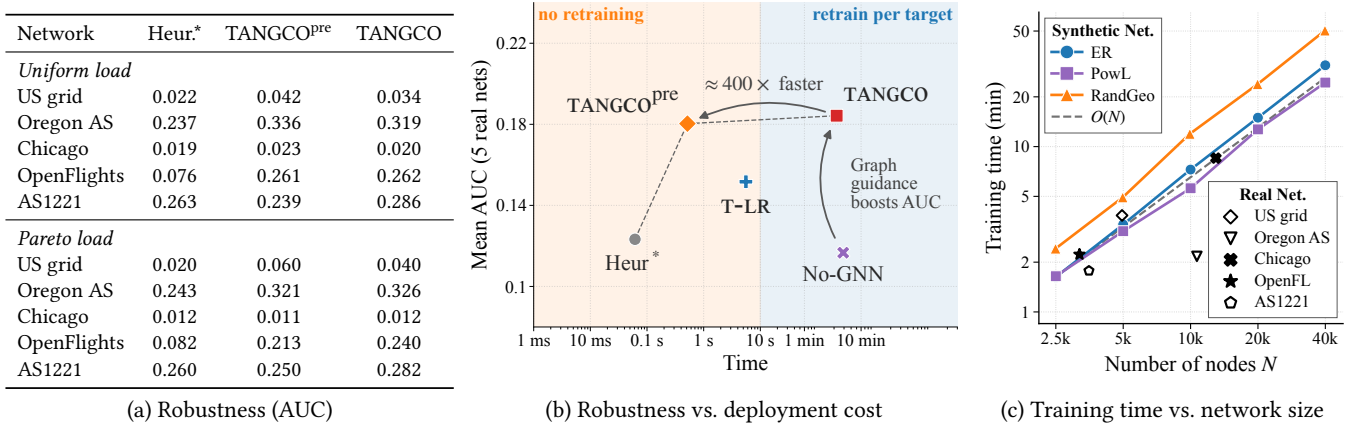

\centering
\resizebox{\textwidth}{!}{\usebox{\teaserrow}}%
\caption{\textbf{\methodpre matches per-network training and lifts mean robustness about 1.5$\times$ over the best heuristic (up to 3$\times$ on hub-heavy networks) on unseen real networks (a), delivers it at heuristic deployment cost (b), and trains in time near-linear in network size (c).} (a)~Absolute robustness (AUC) on five real networks under uniform and Pareto load at $B{=}0.75$; \methodpre is pre-trained on synthetic graphs and applied with no per-network training, while \method trains per network. (b)~Robustness versus time to first allocation on a new network. (c)~Median training time versus number of nodes.}
\label{fig:teaser}
\end{figure*}
\noindent
\textbf{Reproducibility:~} Code and evaluation artifacts will be released upon acceptance.

%% file: sections/formulation.tex
\section{Problem Formulation}
\label{sec:formulation}

We study overload-based cascading failures on an undirected graph
$G=(\mathcal{V},\mathcal{E})$ with $N=|\mathcal{V}|$ nodes.
Each node $v\in\mathcal{V}$ has an initial load $\ell_v$ and a capacity
$c_v$. The capacity is the maximum load that the node can carry before
it fails. We write
$\mathbf{l}=(\ell_v)_{v\in\mathcal{V}}$ and
$\mathbf{c}=(c_v)_{v\in\mathcal{V}}$ for the load and capacity vectors.

\subsection{Cascade Dynamics}
\label{subsec:cascade_dynamics}

Consider a fixed graph $G$, load vector $\mathbf{l}$, capacity vector
$\mathbf{c}$, and initial failure set $\mathcal{F}_0\subseteq\mathcal{V}$.
The cascade starts by removing the nodes in $\mathcal{F}_0$. The loads of these
nodes are then redistributed to the remaining network. After this
redistribution, the surviving nodes are checked for overload. The same
procedure is repeated in discrete iterations $t=1,2,\dots$ until no new
node fails.

Let $\ell_v^{(t)}$ denote the load carried by node $v$ at iteration $t$.
The capacity $c_v$ is fixed during the cascade. A node remains active at
iteration $t$ if its current load does not exceed its capacity, i.e., $\ell_v^{(t)} \le c_v$.
If this condition is violated, the node fails and releases its current
load for redistribution.

Let $\mathcal{F}_t$, $\mathcal{A}_t$, and $\mathcal{N}_t(v)$ denote the nodes that fail at iteration $t$, the nodes still active at that iteration, and the active neighbors of a node $v$.
A failing node $v\in\mathcal{F}_t$ splits its current load $\ell_v^{(t)}$ equally among its active neighbors $\mathcal{N}_t(v)$; if it has none ($\mathcal{N}_t(v)=\emptyset$), the load spreads equally over all active nodes so that the total load is conserved in the system.
Failures within an iteration resolve simultaneously, so an active node $u$ accumulates the shares of its failing neighbors along with its part of the global term $g_t$,
\vspace{-0.1cm}
\begin{equation}
  \ell_u^{(t+1)}
  =
  \ell_u^{(t)}
  +
  \sum_{\substack{v\in\mathcal{F}_t \\ u\in\mathcal{N}_t(v)}}
  \frac{\ell_v^{(t)}}{|\mathcal{N}_t(v)|}
  +
  g_t,
  \qquad
  g_t
  =
  \frac{1}{|\mathcal{A}_t|}
  \sum_{\substack{v\in\mathcal{F}_t \\ \mathcal{N}_t(v)=\varnothing}}
  \ell_v^{(t)}.
  \label{eq:redistribution}
\end{equation} 
This additional load may cause neighboring nodes to exceed their capacities, which can trigger further failures in later iterations.

The cascade stops when an iteration produces no new failures.
Let $t^{*}$ denote the stopping iteration. 
The set of nodes still active at $t^*$ is the \emph{survivor set}~$\mathcal{S}$.
Then, for fixed $G$, $\mathbf{l}$, $\mathbf{c}$, and $\mathcal{F}_0$, the final surviving fraction is
\begin{equation}
  \Phi(G,\mathbf{l},\mathbf{c};\mathcal{F}_0)
  =
  \frac{|\mathcal{S}|}{N}.
\end{equation}
We use this quantity as the robustness measure for a single
cascade.

\textbf{Computational Complexity.~} Each cascade is cheap: a node fails at most once, so redistribution touches each edge once, and each round tests surviving nodes against their capacities.
A cascade halting after $\bar{t}$ rounds costs $O(N\bar{t}+E)$, with $\bar{t}\le N$ in the worst case but far smaller in practice (Section~\ref{sec:experiments:scalability}).

\subsection{Capacity Allocation Problem}
\label{subsec:capacity_allocation}

The cascade model above determines the survival outcome for any fixed capacity assignment.
Next, we consider how to allocate a given total capacity budget across the nodes of a given input graph to maximize robustness.
This requires defining the feasible set of allocations and aggregating the robustness metric across failure sizes into a single objective.

\subsubsection{\textbf{Capacity Constraint}}

Because additional capacity is costly in practice, we fix a total capacity budget and treat the node capacities as the decision variables.
With an unlimited budget, robustness is trivially maximized.

We parameterize a node $v$'s capacity as
\begin{equation}
  c_v = \ell_v + s_v ,
\end{equation}
where $s_v \ge 0$ is the excess capacity, or free space, assigned to node $v$ before any initial failures. 
This parameterization rules out trivial initial failures and simplifies the notation.
Rather than optimize over capacities subject to $\ell_v \leq c_v$, we optimize over non-negative free space under a fixed budget on the total free space.
Let $\mathbf{s}=(s_v)_{v\in\mathcal{V}}$ denote the free-space vector.
For a given free-space budget $B$, the feasible allocations are
\begin{equation}
  \Delta_B
  =
  \left\{
  \mathbf{s}\in\mathbb{R}_+^N:
  \sum_{v\in\mathcal{V}} s_v = B
  \right\}.
  \label{eq:spare_capacity_simplex}
\end{equation}
Thus, optimizing the capacity vector $\mathbf{c}$ is equivalent to choosing a free-space allocation $\mathbf{s}\in\Delta_B$, with capacities given by $c_v=\ell_v+s_v$.

Given $G$, $\mathbf{l}$, $\mathbf{s}$, and an initial failure set $\mathcal{F}_0$, we write the final survival fraction as $\Phi(G,\mathbf{l},\mathbf{s};\mathcal{F}_0)$.

\subsubsection{\textbf{Robustness Objective}}
\label{subsec:robustness_objective}

The fraction $\Phi(G,\mathbf{l},\mathbf{s};\mathcal{F}_0)$ measures robustness against one fixed initial failure set, $\mathcal{F}_0$.
To evaluate an allocation for failures of a given size, let
\begin{equation}
  \mathbb{F}_k
  =
  \left\{
  \mathcal{F}_0\subseteq\mathcal{V}: |\mathcal{F}_0|=k
  \right\}
\end{equation}
be the collection of all initial failure sets containing exactly $k$ nodes.
The exact average survival fraction under $k$ initial failures is
\begin{equation}
  J_k(G,\mathbf{l},\mathbf{s})
  =
  \frac{1}{\binom{N}{k}}
  \sum_{\mathcal{F}_0\in\mathbb{F}_k}
  \Phi(G,\mathbf{l},\mathbf{s};\mathcal{F}_0).
  \label{eq:avg_survival_k}
\end{equation}
This is a finite average over the initial failure sets of size $k$.
Weighting the members of $\mathbb{F}_k$ equally evaluates expected survival under uniformly random initial failures; targeted removal calls for a worst-case rather than an average-case objective and lies outside our scope.
To obtain a single robustness score across failure sizes, we aggregate the average survival fractions over a set of failure sizes $\mathcal{K}\subseteq\{0,\dots,N\}$:
\begin{equation}
  \mathrm{AUC}(G,\mathbf{l},\mathbf{s})
  =
  \sum_{k\in\mathcal{K}}
  w_k J_k(G,\mathbf{l},\mathbf{s}),
  \label{eq:auc}
\end{equation}
where $w_k\ge 0$ and $\sum_{k\in\mathcal{K}} w_k=1$. 
The weights specify how much emphasis is placed on different failure sizes. Plotting $J_k$ against the failure fraction $p=k/N$ traces the survival curve $J(p)$ (Figure~\ref{fig:curves} shows an example), and with uniform weights the objective is the discretized area under that curve, hence the name AUC.
In our experiments we place uniform weight on failure fractions $p\in[0,0.5]$ and zero weight outside this range.
This range focuses the evaluation on low-to-moderate failure fractions, where robustness depends strongly on how the cascade propagates after the initial failures, while keeping the same evaluation criterion for all methods.

In short, $\Phi$ is the basic survival metric for one cascade realization, $J_k$ averages this metric over all initial failure sets of a given size, and AUC aggregates these across the evaluated failure sizes.

Then, given an input graph $G$ and budget $B$, the capacity allocation objective is

\vspace{-0.2in}
\begin{equation}\max_{\mathbf{s}\in\Delta_B}
  \ \mathrm{AUC}(G,\mathbf{l},\mathbf{s}).
  \label{eq:capacity_problem}
\end{equation}

\subsubsection{\textbf{Why the Optimization is Difficult}}
\label{subsec:problem_difficulty}

Although the feasible set $\Delta_B$ is simple, the objective in \eqref{eq:capacity_problem} is not.
Each node's survival is set by a hard overload threshold, so the outcome stays fixed as $\mathbf{s}$ varies until a threshold is crossed, then jumps.
The objective $\mathrm{AUC}(G,\mathbf{l},\mathbf{s})$ is therefore step-like and non-concave: convex methods do not apply, and gradients are uninformative on flat regions and undefined at the jumps.

Exact evaluation is also intractable: $J_k(G,\mathbf{l},\mathbf{s})$ sums over $\binom{N}{k}$ initial failure sets, each requiring a full multi-round cascade.
Because the outcome couples topology, load and capacity, and the location and order of failures, no closed-form characterization is available as in prior global-redistribution models~\cite{zhang_optimizing,ozel2018uniform,irsoy_optimization_multiplex}, which motivates a simulation-based learning approach.

%% file: sections/method.tex
\section{Proposed \method}
\label{sec:method}

\begin{figure*}[t]
    \centering
    \includegraphics[width=0.75\linewidth]{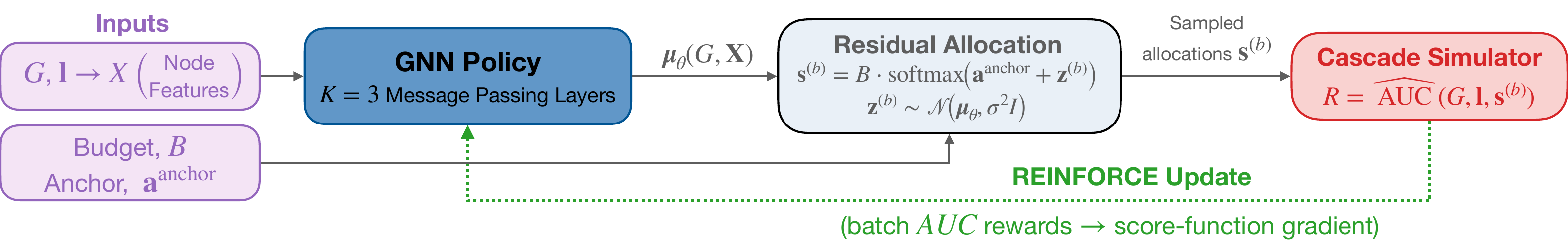}
    \vspace{-10pt}
    \caption{\textbf{\method turns a graph and its loads into a budget-feasible capacity allocation, trained end to end through the cascade simulator.} Node features feed a message-passing policy whose residual output is projected onto the budget simplex; the simulator scores the resulting allocation, and the AUC reward drives a REINFORCE update.}
    \label{fig:system}
\end{figure*}

\method combines a message-passing GNN policy with simulation-based reinforcement learning.
For a fixed instance $(G,\mathbf{l})$ and free-space budget $B$, the goal is to learn a policy that outputs an allocation $\mathbf{s}\in\Delta_B$ with high robustness.
Figure~\ref{fig:system} summarizes the pipeline: node features feed a GNN policy, whose output is projected onto the budget simplex as a feasible free-space allocation.

The allocation can only be evaluated through the cascade simulator, whose hard failure thresholds and discrete redistribution steps make the robustness score non-differentiable.
We therefore train the policy with REINFORCE~\cite{williams1992}: at each iteration, the algorithm samples candidate allocations, scores them using empirical AUC rewards, and updates the policy from these rewards.
The next subsections describe each component.

\subsection{Node Features and GNN Policy}
\label{subsec:gnn_policy}

The allocation depends on both the load vector and each node's local topology.
We therefore use a message-passing GNN rather than learning \(N\) unrelated parameters.
The GNN shares parameters across nodes and aggregates multi-hop information, which is useful since failures can propagate beyond one round.

For each node \(v\) we compute a log-transformed feature vector \(\mathbf{x}_v\) from its own load, its degree, the mean, maximum, and variance of its one-hop neighborhood loads, and its two-hop degree (Appendix~\ref{app:features} gives the formulas).
The GNN stacks \(K\) message-passing layers; its architecture follows GraphSAGE~\cite{hamilton2017inductive} with mean aggregation, differing in two respects: messages are normalized by the sender's degree, in parallel with the load-redistribution mechanism where a failing node splits its load among its neighbors, and each layer adds a residual connection to its MLP update to avoid oversmoothing.
A final linear map produces one scalar residual logit \(\mu_{\theta,v}\) per node, and we write \(\boldsymbol{\mu}_{\theta}(G,\mathbf{X})=(\mu_{\theta,v})_{v\in\mathcal{V}}\) for the vector of outputs (Appendix~\ref{app:gnn} gives the full equations).
The output layer is zero-initialized, so the initial residual logits are exactly zero and the policy starts at the anchor allocation described next.

\subsection{Residual Softmax Allocation}
\label{subsec:residual_softmax}

Directly searching over all allocations in \(\Delta_B\) is high-dimensional and noisy.
We therefore learn a residual correction around a fixed anchor allocation: the anchor gives the policy a reasonable starting point, and the GNN shifts capacity from it using the graph and load features.

Let \(\mathbf{s}^{\mathrm{anchor}}\in\Delta_B\) be a fixed anchor allocation.
We use two anchors in the experiments.
The first is the uniform allocation,
\begin{equation}
  s_v^{\mathrm{uniform}}
  =
  \frac{B}{N}.
\end{equation}
Beyond serving as an intuitive baseline, the uniform allocation is optimal under global redistribution, where the network is treated as fully connected~\cite{zhang_optimizing}, with extensions to partial load loss~\cite{ozel2018uniform} and max-load targeted attacks~\cite{ozel_max_load_attack}.

The second anchor is a local-redistribution heuristic. Irsoy and Ya\u{g}an~\cite{irsoy_decoupled} allocate free space in proportion to a node's one-hop local-risk, giving the Localized Risk-based Free-Space Allocation (LR-FSA) heuristic,
\begin{equation}
\label{eq:rv}
  s_v^{\mathrm{lrfsa}}
  =
  B
  \frac{r_v}{\sum_{u\in\mathcal{V}} r_u},\qquad
  r_v
  =
  \sum_{u\in\mathcal{N}(v)}
  \frac{\ell_u}{d_u}.
\end{equation}
The score \(r_v\) estimates the load that node \(v\) may receive from its neighbors in one round of local redistribution. We train a separate policy for each anchor.
Reported \method performance uses the better of the two completed runs (each trained and evaluated independently), a two-start procedure that uses the cascade simulator already required for training.

We encode an anchor allocation as a logit vector \(\mathbf{a}^{\mathrm{anchor}}\) by taking elementwise logs,
\begin{equation}
 \mathbf{a}^{\mathrm{anchor}} = \log\!\left(\mathbf{s}^{\mathrm{anchor}} + \epsilon\right).
  \label{eq:anchor_logits}
\end{equation}
Since \(\sum_v s_v^{\mathrm{anchor}} = B\), the softmax in Eq.~\eqref{eq:softmax_allocation} maps these logits back to \(\mathbf{s}^{\mathrm{anchor}}\) at initialization; the budget \(B\) re-enters through the leading factor in Eq.~\eqref{eq:softmax_allocation}, and the small floor \(\epsilon\) keeps the log finite where an anchor entry is zero.
During training, the GNN output \(\boldsymbol{\mu}_{\theta}(G,\mathbf{X})\) is treated as the mean of a Gaussian over residual logits.
We sample residual logits as
\begin{equation}
  \mathbf{z}^{(b)}
  \sim
  \mathcal{N}
  \left(
  \boldsymbol{\mu}_{\theta}(G,\mathbf{X}),
  \sigma^2 I
  \right),
  \label{eq:sampled_residual_logits}
\end{equation}
where \(\sigma>0\) controls exploration and is annealed downward over training, and add the residual to the anchor logits before a softmax that projects onto the budget simplex:
\begin{equation}
  s_v^{(b)}
  =
  B
  \frac{
  \exp\left(a_v^{\mathrm{anchor}}+z_v^{(b)}\right)
  }{
  \sum_{u\in\mathcal{V}}
  \exp\left(a_u^{\mathrm{anchor}}+z_u^{(b)}\right)
  },
  \qquad v\in\mathcal{V}.
  \label{eq:softmax_allocation}
\end{equation}
By construction, \(\mathbf{s}^{(b)}\in\Delta_B\) for every sample.
At initialization \(\boldsymbol{\mu}_{\theta}=0\) (Section~\ref{subsec:gnn_policy}), so the deterministic policy with \(\sigma=0\) exactly recovers the anchor, while stochastic samples are centered around it in logit space; as training progresses the GNN learns residual adjustments that improve robustness.

The residual parameterization gives REINFORCE a warm start and reduces training noise, at the cost of restricting the search to softmax perturbations of the chosen anchor.
This is not binding in our experiments, where the best allocation builds on one of the two anchors in every tested case, but it motivates future work on improved anchors and less anchor-tied parameterizations.

\subsection{Simulator-Based Reward}
\label{subsec:simulator_reward}

Each sampled allocation is scored by the cascade model of Section~\ref{subsec:cascade_dynamics}.
Exact AUC evaluation is infeasible, so we use a Monte Carlo estimate: over a finite grid \(\mathcal{P}\) of failure fractions, we draw \(M\) initial failure sets \(\mathcal{F}_{p,1},\dots,\mathcal{F}_{p,M}\) of size \(\lfloor pN\rfloor\) at each \(p\in\mathcal{P}\), giving the empirical survival fraction
\begin{equation}
  \widehat{J}_p(G,\mathbf{l},\mathbf{s})
  =
  \frac{1}{M}
  \sum_{m=1}^{M}
  \Phi(G,\mathbf{l},\mathbf{s};\mathcal{F}_{p,m}),
  \label{eq:empirical_survival}
\end{equation}
and the reward is the empirical AUC over the grid,
\begin{equation}
  R(\mathbf{s})
  =
  \widehat{\mathrm{AUC}}(G,\mathbf{l},\mathbf{s})
  =
  \frac{1}{|\mathcal{P}|}
  \sum_{p\in\mathcal{P}}
  \widehat{J}_p(G,\mathbf{l},\mathbf{s}),
  \label{eq:empirical_auc}
\end{equation}
the empirical counterpart of~\eqref{eq:auc} with uniform weights.
Appendix~\ref{app:training} details how the grid \(\mathcal{P}\) is chosen and how failure sets are pre-sampled and shared across allocations.

\subsection{Non-differentiable Optimization}
\label{subsec:reinforce_training}

The cascade simulator contains threshold failures and discrete rounds, so we cannot differentiate through it.
We instead maximize the expected empirical AUC under the stochastic residual-logit policy,
\begin{equation}
  \mathcal{J}(\theta)
  =
  \mathbb{E}_{\mathbf{z}\sim p_{\theta}}
  \left[
  \widehat{\mathrm{AUC}}
  \left(
  G,\mathbf{l},\mathbf{s}(\mathbf{z})
  \right)
  \right],
  \label{eq:policy_objective}
\end{equation}
for a fixed instance \((G,\mathbf{l})\), budget \(B\), and anchor, where \(p_{\theta}\) is the Gaussian in~\eqref{eq:sampled_residual_logits} and \(\mathbf{s}(\mathbf{z})\) the softmax allocation in~\eqref{eq:softmax_allocation}.
We optimize it with REINFORCE: the score-function estimator turns the simulator's scalar reward into a gradient on the GNN parameters through the log-probability of the sampled logits, and we reduce its variance with a batch-mean baseline over the \(Q\) samples per iteration and stabilize it with a small L2 penalty on the residual logits (Appendix~\ref{app:training}).
We train with Adam~\cite{kingma2015adam}, periodically evaluate the deterministic policy (\(\sigma=0\)) on a held-out set of failure samples, and keep the allocation with the best validation AUC as the final output for that instance.

%% file: sections/experiments.tex
\section{Experiments}
\label{sec:experiments}

We evaluate \method through the following research questions (RQs):
\begin{itemize}
[leftmargin=10pt, itemsep=0.005in, topsep=0pt]
    \item \textbf{RQ1: Effectiveness} Does \method improve robustness over the baseline heuristics?
    \item \textbf{RQ2: Transferability} Does a trained policy transfer to held-out instances within a family, and across families?
    \item \textbf{RQ3: Scalability} How does \method's allocation optimization grow with network size?
    \item \textbf{RQ4: Gain Attribution} Does the gain come from message-passing inductive bias or black-box numerical optimization?
    \item \textbf{RQ5: Gain Characterization} Which topologies, load distributions, and budgets yield the largest robustness gains?
\end{itemize}

\begin{figure}[t]
    \centering
    \includegraphics[width=\columnwidth]{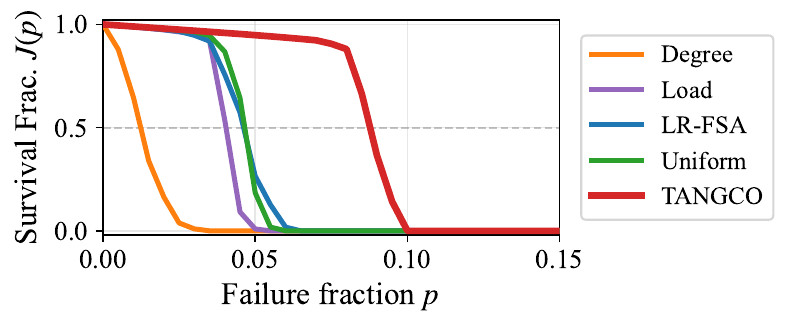}
    \vspace{-20pt}
    \caption{\textbf{\method sustains survival to roughly twice the failure fraction of the best heuristic.} Each curve plots the survival fraction $J(p)$ against failure fraction $p$ for the four heuristics and \method on \textsc{ClusPowL} under uniform load at $B=0.5$ (single realization); larger area under the curve (AUC) is better.}
    \vspace{-10pt}
    \label{fig:curves}
\end{figure}

\subsection{Experimental Setup}
\label{sec:experiments:setup}

\paragraph{Graphs.}
We use five synthetic families and five real-world networks spanning power, road, air-transport, and Internet router/AS domains.
Each synthetic family has $N=5000$ nodes and about $12$ edges per node, and captures a distinct structural regime: \textsc{ER} (homogeneous degrees, no hubs), \textsc{PowL} (a heavy-tailed degree distribution, exponent $\gamma=2.5$), \textsc{ClusPowL} (the same degrees with local clustering $C_c\approx0.15$), \textsc{CorPer} (a dense core with a sparse periphery), and \textsc{RandGeo} (spatial proximity edges with high clustering).
The five real networks span few infrastructure domains: the Western US power grid ($N=4941$), the Oregon AS graph ($N=10{,}670$), the Chicago road network ($N=12{,}979$), the OpenFlights airport network ($N=3188$), and the Rocketfuel AS1221 topology ($N=3515$).
They cover a diverse set of structures, from near-planar (Chicago, maximum degree $7$) to hub-heavy (OpenFlights and Oregon, maximum degree $248$ and $2312$).
Appendix~\ref{app:graphs} gives generation procedures, citations, and per-graph statistics.
For each synthetic family we generate ten independent realizations and report the mean over the ten.

\paragraph{Loads and budgets.}
We evaluate each graph under three initial-load distributions: \emph{uniform} (bounded variation around the mean), \emph{Pareto} (a heavy tail, so a few nodes carry much larger loads), and \emph{bimodal} (a two-group high/low mixture).
The budget $B$ is the total free space to allocate; we set $B/\sum_v\ell_v\in\{0.5,0.75,1.0\}$, from tight to loose, giving nine load-budget conditions in all.

\paragraph{Baseline heuristics.}
We compare \method with four deterministic capacity rules from the cascade literature, all normalized to $\sum_{v}s_v=B$: the two anchors of Section~\ref{subsec:residual_softmax}, \textsc{Uniform} (equal free space) and \textsc{LR-FSA} (one-hop local-risk); \textsc{Load}, proportional to initial load~\cite{MotterLai2002}; and \textsc{Degree}, proportional to degree~\cite{li2008limited}.

\begin{figure*}[t]
    \centering
    \includegraphics[width=0.9\linewidth]{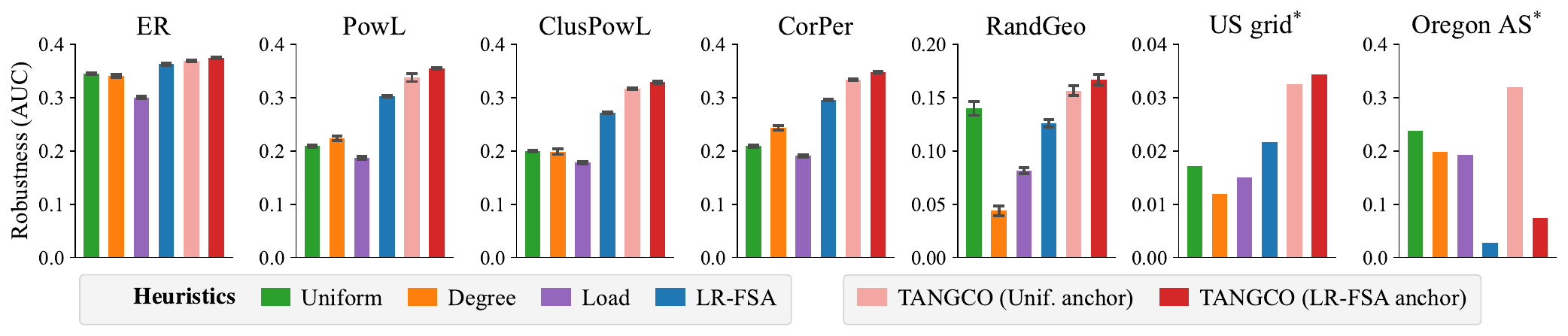}
    \vspace{-10pt}
    \caption{\textbf{\method outperforms all baselines across synthetic and real topologies.} Absolute robustness (AUC) of four heuristics and \method w/ two anchored policies on five synthetic families and two real networks at uniform load and $B=0.75$. Error bars span one standard deviation over 10 realizations; the starred networks (US grid, Oregon AS) are single real graphs and carry no error bars. See Table~\ref{tab:primary} for all real networks across all nine load-budget conditions.}
    \label{fig:headtohead}
\end{figure*}

\paragraph{Architecture, training, and protocol.}
We use $K=3$ message-passing layers and hidden width $64$ throughout, except in the depth ablation of Section~\ref{sec:experiments:depth}, and train each instance with Adam~\cite{kingma2015adam} (Appendix~\ref{app:training} lists all hyperparameters).
We report the robustness score $\mathrm{AUC}(G,\mathbf{l},\mathbf{s})$ of Eq.~\eqref{eq:auc}, the average survival fraction over $p\in[0,0.5]$.
The \emph{primary} comparison is \method (maximum AUC over the two anchors) against the best heuristic (maximum AUC over \textsc{Uniform}, \textsc{Degree}, \textsc{Load}, \textsc{LR-FSA}) on each configuration.
We replicate training over five seeds per configuration and report the mean; training-seed variance is small (Appendix~\ref{app:training}).

\subsection{Overall Performance (RQ1)}
\label{sec:experiments:main}

Figure~\ref{fig:curves} shows a robustness curve for a representative instance and makes the AUC metric concrete: each curve traces the survival fraction as the failure fraction grows, and a more robust allocation keeps the curve higher for longer, enclosing more area. \method pushes its curve well to the right of every heuristic, holding a high survival fraction out to roughly twice the failure fraction that the heuristics withstand.

Figure~\ref{fig:headtohead} compares the absolute AUC of the four heuristics and the two anchored policies at a representative configuration, uniform load and $B=0.75$.
Table~\ref{tab:primary} reports the relative improvement of \method over the best heuristic across all nine load-budget conditions.
Two observations follow.

\begin{itemize}
[leftmargin=10pt, itemsep=0.025in, topsep=0pt]
\item \textbf{\method improves on the best heuristic in almost every condition.}
Figure~\ref{fig:headtohead} shows the comparison on seven representative networks at uniform load and $B=0.75$, where \method exceeds every heuristic.
Table~\ref{tab:primary} extends it to the full set: the policy improves on the best heuristic in all $450$ synthetic cells ($5$ families $\times$ $9$ conditions $\times$ $10$ realizations) and in $40$ of the $45$ real-world conditions, with gains from $+1.6\%$ on \textsc{ER} at a loose budget to $+246\%$ on \textsc{OpenFlights}.
The five exceptions all sit at the tightest budget on a spatially embedded network (Chicago under all three loads, the US grid under uniform and bimodal loads), where the improvement is $\approx0$ because no allocation survives even $p\approx0.01$ (Section~\ref{sec:experiments:regime}).
\item \textbf{Training two anchors guards against anchor failure.}
We anchor the policy on either \textsc{Uniform} or \textsc{LR-FSA} and keep the better of the two.
\textsc{LR-FSA} is the stronger prior in most conditions, but it risks trapping the policy: in Figure~\ref{fig:headtohead}, the \textsc{LR-FSA} anchor on Oregon AS collapses along with the heuristic it starts from, which concentrates too much free space on a few extremely connected hubs.
The \textsc{Uniform} anchor is the more stable prior, supplying the winning policy on both hub-heavy AS topologies, at the risk of missing the gain \textsc{LR-FSA} offers where its prior holds.
Keeping the better of the two captures \textsc{LR-FSA}'s upside without its risk; Appendix~\ref{app:anchor} reports the per-family anchor win-counts and illustrates the collapse on Oregon AS.
\end{itemize}

\begin{table}[b]
\centering
\vspace{-10pt}
\caption{\method achieves positive relative AUC improvements over the best heuristic across all 450 synthetic cells and 40 of 45 real-world conditions.
Synthetic entries are means over ten realizations, each averaged over five seeds; real-world entries are means over five training seeds.}
\vspace{-0.1in}
\label{tab:primary}
\setlength{\tabcolsep}{2.5pt}
\footnotesize
\begin{tabular}{llrrrrr}
\toprule
 & & \multicolumn{5}{c}{Synthetic families} \\
\cmidrule(l){3-7}
Load & Budget & \textsc{ER} & \textsc{PowL} & \textsc{ClusPowL} & \textsc{CorPer} & \textsc{RandGeo} \\
\midrule
Uniform & $B{=}0.5$  & $+8.0$ & $+54.9$ & $+76.7$ & $+46.1$ & $+17.3$ \\
        & $B{=}0.75$ & $+3.3$ & $+17.2$ & $+21.2$ & $+17.4$ & $+19.3$ \\
        & $B{=}1.0$  & $+1.6$ & $+8.2$ & $+10.0$ & $+8.5$ & $+15.1$ \\
\midrule
Pareto  & $B{=}0.5$  & $+18.1$ & $+64.9$ & $+85.0$ & $+61.6$ & $+19.7$ \\
        & $B{=}0.75$ & $+5.7$ & $+19.3$ & $+25.1$ & $+20.2$ & $+44.8$ \\
        & $B{=}1.0$  & $+2.7$ & $+9.2$ & $+11.3$ & $+9.4$ & $+22.0$ \\
\midrule
Bimodal & $B{=}0.5$  & $+13.0$ & $+60.6$ & $+89.4$ & $+50.0$ & $+23.8$ \\
        & $B{=}0.75$ & $+4.5$ & $+19.4$ & $+23.8$ & $+19.3$ & $+37.3$ \\
        & $B{=}1.0$  & $+2.1$ & $+9.2$ & $+11.0$ & $+10.0$ & $+20.0$ \\
\midrule
 & & \multicolumn{5}{c}{Real-world networks} \\
\cmidrule(l){3-7}
Load & Budget & US grid & Oregon AS & Chicago & OpenFl. & AS1221 \\
\midrule
Uniform & $B{=}0.5$  & $\approx 0$ & $+67.2$ & $\approx 0$ & $+245.8$ & $+22.6$ \\
        & $B{=}0.75$ & $+58.6$ & $+34.4$ & $+4.9$ & $+244.9$ & $+9.0$ \\
        & $B{=}1.0$  & $+38.4$ & $+22.9$ & $+10.2$ & $+128.1$ & $+2.5$ \\
\midrule
Pareto  & $B{=}0.5$  & $+3.0$ & $+107.5$ & $\approx 0$ & $+231.3$ & $+20.3$ \\
        & $B{=}0.75$ & $+97.0$ & $+34.2$ & $+4.1$ & $+193.3$ & $+8.5$ \\
        & $B{=}1.0$  & $+28.3$ & $+21.6$ & $+3.9$ & $+123.1$ & $+6.2$ \\
\midrule
Bimodal & $B{=}0.5$  & $\approx 0$ & $+66.6$ & $\approx 0$ & $+204.5$ & $+13.9$ \\
        & $B{=}0.75$ & $+60.4$ & $+34.1$ & $+5.1$ & $+204.2$ & $+9.3$ \\
        & $B{=}1.0$  & $+20.5$ & $+18.9$ & $+10.6$ & $+142.0$ & $+5.4$ \\
\bottomrule
\end{tabular}
\end{table}

\subsection{Transferability (RQ2)}
\label{sec:experiments:transfer}

Every result so far trains one policy per graph instance.
We now ask whether a trained policy transfers to unseen graphs: first across held-out instances within and between families, then as a model deployable on any new network.

\textbf{Per-graph policies transfer within a family and across families that share structure.}
For within-family transfer we train one policy jointly on eight of a family's ten realizations and evaluate it on the two held out; for cross-family transfer we apply each such policy to every other family's held-out graphs.
Table~\ref{tab:transfer} reports the $5\times5$ source-to-target matrix of improvement in AUC over the best heuristic (uniform load, $B=0.75$), with per-instance training in the bottom row.
Within-family transfer is positive on all five families and matches per-instance training within about $\pm0.002$ AUC.
Cross-family, transfer still improves on the target heuristic in $17$ of $20$ cells, holding among the heavy-tailed families \textsc{PowL}, \textsc{ClusPowL}, and \textsc{CorPer}, and failing only when a \textsc{RandGeo} policy, trained on spatial graphs without hubs, is applied to a hub-dominated target (e.g.\ \textsc{RandGeo}$\to$\textsc{PowL}, $-0.062$).
Pareto load shows the same pattern, improving in $14$ of $20$ cross-family cells (Appendix~\ref{app:transfer}).

\begin{table}[t]
\centering
\caption{A policy trained on one family matches instance-specific \method on held-out graphs from the same family and transfers across structurally similar families.
Entries show improvement in AUC over the best heuristic on the target under uniform load and $B=0.75$, reporting the better of the two training anchors (per-anchor values in Appendix~\ref{app:transfer}).
Diagonal cells show within-family transfer from eight training realizations to two held-out realizations; off-diagonal cells show cross-family transfer.
The bottom row reports \method gains on the same targets.
Shading marks the \textsc{PowL}, \textsc{ClusPowL}, and \textsc{CorPer} block, where policies transfer with little loss.}
\label{tab:transfer}
\vspace{-0.15in}
\small
\begin{tabular}{lrrrrr}
\toprule
 & \multicolumn{5}{c}{Target} \\
\cmidrule(l){2-6}
Source & \textsc{ER} & \textsc{PowL} & \textsc{ClusPowL} & \textsc{CorPer} & \textsc{RandGeo} \\
\midrule
\textsc{ER}       & \textbf{+.011} & +.005          & +.015          & +.027          & +.014 \\
\textsc{PowL}     & +.012          & \cellcolor{gray!15}\textbf{+.053} & \cellcolor{gray!15}+.057 & \cellcolor{gray!15}+.047 & +.021 \\
\textsc{ClusPowL} & +.011          & \cellcolor{gray!15}+.050 & \cellcolor{gray!15}\textbf{+.056} & \cellcolor{gray!15}+.045 & +.023 \\
\textsc{CorPer}   & +.012          & \cellcolor{gray!15}+.043 & \cellcolor{gray!15}+.049 & \cellcolor{gray!15}\textbf{+.051} & +.023 \\
\textsc{RandGeo}  & +.008          & $-$.062        & $-$.044        & $-$.017        & \textbf{+.036} \\
\midrule
\method      & +.011          & +.051          & +.058          & +.050          & +.026 \\
\bottomrule
\end{tabular}
\end{table}

\textbf{\methodpre transfers to unseen real networks with no per-target training.}
The cross-family result suggests that a policy exposed to enough structural variety during training should transfer broadly.
We therefore pre-train \methodpre on a curated suite of $64$ synthetic graphs spanning different families and a wide range of sizes, densities, and family-specific parameters (separate checkpoint per anchor; Appendix~\ref{app:graphs}).
Applied with no parameter update on the target (reporting the better of the two anchors), \methodpre reaches mean AUC $0.18$ on the five real networks under uniform load, matching per-instance training and improving on the best heuristic ($0.12$) by about 1.5$\times$ on average, up to 3$\times$ on OpenFlights.
It recovers the gain even where family-specific transfer fails, reaching $0.261$ on OpenFlights against a heuristic $0.076$ and $0.336$ on Oregon AS against $0.237$ (Figure~\ref{fig:teaser}(a)).
It improves on the best heuristic on four of the five real networks under uniform load, and delivers this at heuristic deployment cost (Section~\ref{sec:experiments:scalability}).

%% file: sections/scalability.tex
\subsection{Scalability (RQ3)}
\label{sec:experiments:scalability}

\begin{itemize}
[leftmargin=10pt, itemsep=0.025in, topsep=0pt]
\item \textbf{Training scales near-linearly with network size, and cascade simulator dominates while the GNN stays under $2\%$.}
Each run performs $\approx8.4$M cascade simulations, each $O(N\bar{t}+E)$, while the policy adds one forward and one backward pass per iteration at $O(N+E)$, negligible beside it (Section~\ref{subsec:cascade_dynamics}).
Across three synthetic families up to $40{,}000$ nodes the fitted exponents lie between $0.97$ and $1.10$, and the five evaluated networks fall on the same trend (Figure~\ref{fig:teaser}(c)); \textsc{RandGeo} scales worst, consistent with its deeper cascades.
\item \textbf{Once trained, \methodpre needs no per-target training and deploys at heuristic-scale cost, capturing most of the robustness gain.}
At deployment each pretrained anchor allocates in one forward pass (reported robustness uses the better of the two); wall-clock placement is about half a second per pass, comparable to a heuristic and roughly $400\times$ faster than training from scratch (Figure~\ref{fig:teaser}(b)).
It reaches mean AUC $0.18$ across the five real networks, matching instance-trained \method ($0.18$) and well above the best heuristic ($0.12$).
\tlr, the tuned local-risk rule we introduce in \S\ref{subsec:interp_rule}, improves on the heuristics for a few seconds of search, while per-network training buys the final increment of robustness for minutes.
\end{itemize}

Appendix~\ref{app:scaling} reports the full wall-clock times, the four-panel deployment comparison across both load distributions, and the size-scaling result against edge count.

\begin{table}[b]
\vspace{-10pt}
\centering
\caption{\textbf{\method beats the \nognn variant in all $10$ uniform cells; without message passing, the \nognn variant adds little over the best heuristic.} AUC at uniform load and $B=0.75$, single realization per family. \emph{Heuristic*} is the best of the four heuristics. The learned search variants (\nognn and CMA-ES, a derivative-free optimizer; Appendix~\ref{app:ablation}) are means over five training seeds (best of two anchors). Parenthesised values are each column's shortfall relative to \method.}
\label{tab:ablation}
\vspace{-0.15in}
\footnotesize
\setlength{\tabcolsep}{4pt}
\begin{tabular}{lrrrr}
\toprule
Graph & \method & \nognn & CMA-ES & Heuristic* \\
\midrule
\textsc{ER}       & $0.372$ & $0.358$ ($-3.6\%$)  & $0.357$ ($-3.9\%$)  & $0.357$ ($-3.9\%$) \\
\textsc{PowL}     & $0.354$ & $0.311$ ($-12.2\%$) & $0.313$ ($-11.6\%$) & $0.302$ ($-14.6\%$) \\
\textsc{ClusPowL} & $0.327$ & $0.279$ ($-14.6\%$) & $0.282$ ($-13.6\%$) & $0.270$ ($-17.4\%$) \\
\textsc{CorPer}   & $0.345$ & $0.299$ ($-13.4\%$) & $0.295$ ($-14.5\%$) & $0.295$ ($-14.5\%$) \\
\textsc{RandGeo}  & $0.171$ & $0.144$ ($-16.0\%$) & $0.141$ ($-17.4\%$) & $0.141$ ($-17.4\%$) \\
US grid           & $0.034$ & $0.024$ ($-30.3\%$) & $0.022$ ($-34.8\%$) & $0.022$ ($-36.9\%$) \\
Oregon AS         & $0.319$ & $0.244$ ($-23.4\%$) & $0.259$ ($-18.9\%$) & $0.237$ ($-25.6\%$) \\
Chicago           & $0.020$ & $0.019$ ($-3.8\%$)  & $0.019$ ($-4.6\%$)  & $0.019$ ($-4.7\%$) \\
OpenFlights       & $0.262$ & $0.083$ ($-68.2\%$) & $0.145$ ($-44.6\%$) & $0.076$ ($-71.0\%$) \\
AS1221            & $0.286$ & $0.213$ ($-25.6\%$) & $0.235$ ($-17.9\%$) & $0.263$ ($-8.3\%$) \\
\bottomrule
\end{tabular}
\end{table}

%% file: sections/interpretability.tex
\subsection{Ablations: Optimizer and Depth (RQ4)}
\label{sec:experiments:ablation}
\label{sec:experiments:depth}

A learned policy could beat heuristics by searching the allocation space more
thoroughly than any fixed formula, or by encoding graph structure through message
passing. We isolate the first with two variants that optimize a free allocation
vector directly, \nognn (same REINFORCE protocol as \method) and a derivative-free
CMA-ES optimizer, and the second by varying message-passing depth
(Appendix~\ref{app:ablation},~\ref{app:depth_anchor}).

\textbf{Direct search stays near the best heuristic whether
policy-gradient or evolutionary; message passing carries the gain.}
\nognn barely clears the best heuristic (mean $\Delta$AUC $-0.001$ uniform, $+0.002$ Pareto); CMA-ES gains more ($+0.009$, $+0.012$), mostly on hub-heavy graphs, reaching $0.145$ on \textsc{OpenFlights} against a heuristic $0.076$, but still falls well short of \method ($0.262$). Both lose on AS1221, where \textsc{Degree} is already strong.
\method beats the \nognn variant in all $10$ uniform cells (mean $+0.052$ AUC; Table~\ref{tab:ablation}) and in all $10$ Pareto cells, and beats the CMA-ES variant in every cell (Appendix~\ref{app:ablation}).

\textbf{One message-passing hop is enough.}
A per-node MLP with no message passing ($K=0$) already improves on the best heuristic; the first hop adds further gain, after which $K=2$--$5$ stay flat (Appendix~\ref{app:depth_anchor}, Table~\ref{tab:depth_anchor}).
The $K{=}0\!\to\!1$ step is small under the \textsc{LR-FSA} anchor, which already injects a one-hop local-risk estimate.

\begin{table}[b]
\centering
\caption{\textbf{\tlr recovers most of the heuristic-to-\method gap where local-risk dominates (regime A), and \method matches or beats it on every network.} AUC at uniform load, $B=0.75$ (synthetics: mean over ten realizations; cf.\ Table~\ref{tab:ablation}'s one-realization cut). Heuristic$^{*}$ is the best of the four heuristics; \tlr is \textsc{LR-FSA} with a per-graph exponent fit on training failures.}
\vspace{-0.15in}
\label{tab:a3}
\small
\setlength{\tabcolsep}{4pt}
\begin{tabular}{lrrrrr}
\toprule
\textit{Synthetic:} & \textsc{ER} & \textsc{PowL} & \textsc{ClusPowL} & \textsc{CorPer} & \textsc{RandGeo} \\
\midrule
Heuristic$^{*}$ & 0.362 & 0.303 & 0.273 & 0.297 & 0.132 \\
\tlr            & 0.372 & \textbf{0.348} & \textbf{0.322} & \textbf{0.342} & 0.145 \\
\method         & \textbf{0.374} & \textbf{0.355} & \textbf{0.330} & \textbf{0.347} & \textbf{0.158} \\
\midrule
\textit{Real:} & US grid & OregAS & Chicago & OpenFL & AS1221 \\
\midrule
Heuristic$^{*}$ & 0.022 & 0.237 & 0.019 & 0.076 & 0.263 \\
\tlr            & 0.023 & 0.214 & \textbf{0.019} & \textbf{0.265} & 0.238 \\
\method         & \textbf{0.034} & \textbf{0.319} & \textbf{0.020} & \textbf{0.262} & \textbf{0.286} \\
\bottomrule
\end{tabular}
\end{table}

\subsection{What Drives the Gains (RQ5)}
\label{sec:experiments:regime}
\label{sec:interpretability}
\method's gains concentrate on heterogeneous topologies at intermediate budgets.
Analyzing the learned allocations shows local risk is a strong signal in some regimes but incomplete in others, where \method draws on additional structural and nonlinear signals.

\begin{itemize}
[leftmargin=10pt, itemsep=0.025in, topsep=0pt]
\item \textbf{Gains scale with structural heterogeneity and peak at intermediate budgets.}
The largest improvements fall on heavy-tailed and hub-heavy topologies (\textsc{ClusPowL} $+89\%$, \textsc{PowL} $+65\%$, \textsc{OpenFlights} up to $+246\%$, Oregon AS up to $+108\%$), while homogeneous \textsc{ER} stays between $+1.6\%$ and $+18.1\%$.
Across budgets the gain is largest where the baseline is weak but survivable: it shrinks with budget on the strong-baseline families and grows with budget on the weak-baseline ones.
Chicago and the US grid are the extreme, with no gain at $B{=}0.5$ until the budget buys enough survival to reallocate (Table~\ref{tab:primary}).
\item \textbf{A systematic, nested analysis identifies local risk as the primary signal the policy uses, and where it falls short.}
\label{subsec:interp_signal}
We fit the normalized allocation $\log\!\left(s_v/(B/N)\right)$ with a nested model sequence: the one-hop local-risk score $r_v$ (Eq.~\eqref{eq:rv}) alone, then cumulatively adding degree, load, and $k$-core, then a generalized additive model where the linear fit fails (Appendix~\ref{app:regime}).
Local risk alone explains the allocation in regime~A ($R^2\ge0.90$: the moderately heavy-tailed families \textsc{PowL}, \textsc{ClusPowL}, \textsc{CorPer}, and several networks under uniform load).
The rest need either an additional local feature such as degree (regime~B, e.g.\ \textsc{RandGeo}) or nonlinear effects (regime~C), as on the extreme-hub Oregon AS and AS1221, where a nonlinear fit reaches $0.82$--$0.98$ against linear $0.03$--$0.33$.
Local risk is thus strong but incomplete: \method rescales capacity-versus-risk where it dominates and draws on higher-order structure elsewhere.
\item \textbf{Tuned Local-Risk (\tlr) recovers most of the heuristic-to-\method gap where local risk dominates.}
\label{subsec:interp_rule}
Regime~A motivates fitting a per-graph exponent, $s_v\propto r_v^{\gamma}$, by a bounded search on training failures rather than fixing $\gamma=1$.
\tlr recovers about $85\%$ of the gap in regime~A and matches \method on \textsc{OpenFlights} ($0.265$ vs.\ $0.262$; Table~\ref{tab:a3}).
Outside regime~A it falls below the best heuristic on Oregon AS and AS1221, where only \method's higher-order graph reasoning recovers the gain.
\end{itemize}

%% file: sections/related_work.tex
\section{Related Work}
\label{sec:related}


\paragraph{Cascading failures.}
Cascading failures in flow networks are a long-studied problem. The foundational
model of Motter and Lai assigns each node a load equal to its betweenness and a
capacity fixed to a multiple of that load; when a node fails its load
redistributes to others, which may overload them and trigger a
cascade~\cite{MotterLai2002}. A large body of work builds on this setup to study
how cascades propagate under different mechanisms, including interdependent
networks where failures couple across systems~\cite{buldyrev2010}, dynamical
recovery of failed nodes~\cite{majdandzic2014}, and heterogeneous rules for how a
failed node's load spreads to its neighbors~\cite{hou2017heterogeneous}. A
related line engineers robustness by setting capacity as a fixed function of a
local statistic, either a nonlinear function of load~\cite{wang_kim2007} or a
tuned power of node degree under a fixed budget~\cite{li2008limited}. These
models treat capacity as a closed-form rule fit to load or degree, rather than an
allocation optimized against the cascade itself.

Fewer works take capacity allocation as a decision variable to optimize. Researchers optimize the capacity of a power system for robustness against cascades~\cite{zhang_optimizing}; prove that a uniform
redundancy allocation maximizes robustness under random failures and extend the
analysis to max-load targeted attacks~\cite{ozel2018uniform,ozel_max_load_attack};
and derive optimal load-capacity allocations in multiplex networks~\cite{irsoy_optimization_multiplex}. These analyses assume \emph{global}
redistribution, where a failed node's load spreads equally to all surviving
nodes. Global redistribution yields clean, often closed-form optima, but it
discards the network topology, and the resulting allocations do not transfer to
local redistribution, where a node sheds load onto its neighbors and cascade
outcomes depend on multi-round structure. Our baselines draw on both lines: the
\textsc{Uniform} allocation follows the global-redistribution
optimum~\cite{ozel2018uniform,zhang_optimizing,irsoy_optimization_multiplex},
\textsc{LR-FSA} is a one-hop local-risk heuristic~\cite{irsoy_decoupled},
\textsc{Degree} instantiates the degree-weighted rule~\cite{li2008limited}, and
\textsc{Load} the fixed-tolerance rule~\cite{MotterLai2002}.

\textit{Learning for graph problems.}
A separate line learns to solve graph optimization problems. Graph neural networks
with reinforcement learning construct combinatorial solutions node by
node~\cite{Khalil2017}, and learn constrained resource allocations such as wireless
power control~\cite{eisen2020regnn} or resource distribution in observational
science~\cite{cranmer2021unsupervised}. These methods assume a differentiable
objective or a differentiable relaxation~\cite{eisen2020regnn,cranmer2021unsupervised}.
Our fail-or-survive cascade objective is piecewise-constant and non-differentiable,
so we train \method with score-function gradients (REINFORCE~\cite{williams1992}) and
keep the budget exact through a residual-softmax output.

\textit{Learning for cascading failures.}
Closest to our setting is work that learns to predict or mitigate cascades. Like
\method, Mao~et al. train a graph neural network with reinforcement learning
against a cascade simulator, here to surface the most vulnerable nodes of
interdependent urban infrastructure~\cite{mao2023detecting}. Jhun~et al. reinforce
the highest-ranked nodes under a learned avalanche-centrality measure to suppress
nonlocal cascades~\cite{jhun2023nonlocal}; others treat defense as an adversarial
game over discrete targets~\cite{cunningham2023mitigating}, steer spreading through
node interventions~\cite{meirom2021controlling}, or learn a diffusion
surrogate~\cite{xiang2024hyperparametric}. A broader body predicts cascade outcomes
and risk~\cite{bhaila2024cascading,zhu2023physics,zhu2023realtime}, identifies
vulnerable node sets~\cite{xu2026weakset}, and benchmarks these
tasks~\cite{varbella2024powergraph}. All rank, classify, or discretely reinforce a
node set; none outputs a continuous, budget-feasible capacity allocation optimized
end-to-end through the overload cascade.

%% file: sections/conclusion.tex
\section{Conclusion}
\label{sec:conclusion}

We introduced \method, a topology-aware learning framework for capacity allocation under local load redistribution. The framework contributes in two ways. First, it provides a viable method for improving robustness when the cascade objective is non-differentiable and no analytical allocation rule is available. Second, it helps explain the allocation problem itself by identifying when local-risk is sufficient and when additional or nonlinear structure matters. This analysis led to Tuned Local-Risk (\tlr), a rule that improves on the original heuristic in the regimes it can represent, while also showing why a GNN remains necessary in regimes that cannot be reduced to a simple rule. \methodpre, pre-trained on synthetic graphs, transfers to unseen real networks at the deployment cost of a hand-designed heuristic, extending the method to settings without per-network training. \method therefore serves both as an allocation method and as a tool for extracting simpler principles from cascade dynamics.
Future work can extend the framework to domain-specific redistribution models and develop graph-adaptive anchor selection from features alone.

%% file: sections/appendix.tex
\appendix
\section{Reproducibility}
\label{app:repro}
We provide the reproducibility details needed to reproduce the experiments; the implementation, evaluation code, recipe manifests, and verification artifacts will be released upon acceptance.

\subsection{Notation}
\label{app:notation}

Scalars are lowercase italic, vectors lowercase bold, matrices uppercase bold, and sets calligraphic.
Uppercase italic is reserved for cardinalities and fixed counts, and blackboard bold for a family of sets.

\begin{table}[h]
\centering
\small
\caption{Notation used throughout the paper.}
\label{tab:notation}
\begin{tabular}{ll}
\toprule
Symbol & Meaning \\
\midrule
\multicolumn{2}{l}{\emph{Graph and sets}} \\
$G=(\mathcal{V},\mathcal{E})$   & graph, node set, edge set \\
$\mathcal{N}(v)$, $\mathcal{N}_t(v)$ & neighbors of $v$; those still active at round $t$ \\
$\mathcal{F}_0$, $\mathcal{F}_t$ & initial failure set; nodes failing at round $t$ \\
$\mathcal{A}_t$, $\mathcal{S}$   & nodes active at round $t$; survivor set \\
$\mathbb{F}_k$                   & family of initial failure sets of size $k$ \\
$\Delta_B$                       & budget simplex of feasible allocations \\
$\mathcal{K}$, $\mathcal{P}$     & evaluated failure sizes; failure-fraction grid \\
\midrule
\multicolumn{2}{l}{\emph{Counts}} \\
$N$, $E$                         & number of nodes, edges \\
$B$                              & total free-space budget \\
$K$                              & message-passing layers \\
$Q$                         & sampled allocations per iteration\\
$M$                              & failure samples per fraction \\
\midrule
\multicolumn{2}{l}{\emph{Vectors and matrices}} \\
$\mathbf{l}$, $\mathbf{c}$, $\mathbf{s}$ & load, capacity, free-space allocation \\
$\mathbf{X}$, $\mathbf{x}_v$     & node feature matrix; feature vector of node $v$ \\
$\mathbf{h}_v^{(i)}$             & hidden state of node $v$ at layer $i$ \\
$\boldsymbol{\mu}_{\theta}$, $\mathbf{z}^{(b)}$ & GNN output; sampled residual logits \\
\midrule
\multicolumn{2}{l}{\emph{Scalars}} \\
$\ell_v$, $c_v$, $s_v$           & load, capacity, free space at node $v$ \\
$r_v$, $d_v$                     & one-hop local-risk; degree \\
$t$, $\bar{t}$                   & cascade round; mean rounds to halt \\
$k$, $p$                         & failure-set size; failure fraction $k/N$ \\
$\gamma$, $\sigma$, $\lambda$    & damping exponent; exploration scale; L2 penalty \\
\midrule
\multicolumn{2}{l}{\emph{Functions}} \\
$\Phi$, $J_k$, $J(p)$            & survival fraction; its average over $\mathbb{F}_k$; survival curve \\
$\mathrm{AUC}$                   & robustness objective, Eq.~\eqref{eq:auc} \\
$\mathcal{J}(\theta)$, $\mathcal{L}_{\mathrm{RL}}$ & policy objective; training loss \\
\bottomrule
\end{tabular}
\end{table}

\subsection{Network Data}
\label{app:graphs}

We generate synthetic graphs in two settings that serve different purposes.
The evaluation families fix one canonical setting per topology, so that any difference in results reflects structure alone.
The curated training set does the opposite: it varies structure on purpose, so that a single pre-trained policy meets many regimes before it ever sees a test graph.
Table~\ref{tab:networks} summarizes all three groups.

\paragraph{Synthetic evaluation families.}
All five families share $N=5000$ nodes and mean degree $\langle k\rangle\approx12$, and each fixes the one parameter that defines it.
\textsc{ER}~\cite{erdos_renyi1960,newman_random_graphs} connects each pair of nodes independently with equal probability, giving even degrees and no pronounced hubs.
\textsc{PowL}~\cite{newman_random_w_arbitrary_degree_distributions,newman_configuration_model} draws a power-law degree sequence (exponent $\gamma=2.5$) and wires it with the configuration model, so a few nodes act as strong hubs, as in communication, social, and information networks.
\textsc{ClusPowL}~\cite{newman_random_w_clustering} keeps the \textsc{PowL} degree sequence but rewires it to raise the average local clustering to $C\approx0.15$, adding short cycles and redundant local paths; comparing it with \textsc{PowL} isolates the effect of clustering on cascade propagation.
\textsc{CorPer}~\cite{core_periphery_structures} is a degree-corrected core--periphery model with a core fraction of $0.2$: a dense core carries the main cascade pathways and a sparse periphery surrounds it.
\textsc{RandGeo}~\cite{random_geometric_graphs} places nodes in the unit square and joins those within a fixed radius, giving spatial locality, high clustering, and long paths, as in physically constrained networks such as road and grid systems.

\paragraph{Real-world networks.}
The five measured networks test whether the synthetic trends carry over to empirical topologies: the Western US power grid~\cite{data_us_grid,watts_strogatz1998}, the Oregon AS graph~\cite{leskovec2005graphs,snap_oregon1}, the Chicago road network~\cite{netzschleuder}, the OpenFlights airport route network~\cite{opsahl2010,netzschleuder}, and the Rocketfuel AS1221 router topology~\cite{spring2004rocketfuel}.
Together they span the structural range of the synthetic families, from the low-degree, long-path grid and road networks to the extreme-hub AS and airport graphs (Table~\ref{tab:networks}).
OpenFlights carries a domain caveat: equal-split local redistribution is a stylized abstraction for air transport, since a substitutable airport is not always a route neighbor.

\paragraph{Curated training set.}
The pre-trained policy \methodpre never sees the canonical settings above.
We train it on a curated suite of $64$ synthetic graphs built for structural variety alone: $16$ graphs from each of four families (\textsc{ER}, \textsc{PowL}, \textsc{CorPer}, \textsc{RandGeo}), spread across four size bands (roughly $2{,}000$ to $15{,}000$ nodes) and four mean-degree bands (from sparse, $\langle k\rangle\approx2$, to dense, $\langle k\rangle\approx16$).
Within each family we also open up the defining parameter instead of pinning it: the power-law exponent ranges over $\gamma\in[2.1,3.2]$, the core--periphery core fraction over $[0.11,0.45]$, and the random-geometric radius over $[0.011,0.034]$.
\textsc{ER} has no shape parameter, so only its size and density vary.
We leave \textsc{ClusPowL} out of training on purpose.
It is a clustered cousin of \textsc{PowL}, and the heavy-tailed families already transfer well among themselves (Table~\ref{tab:transfer}), so adding it would tilt the four-family balance toward that regime.
Holding it out instead tests whether the policy carries over to a related family it never trained on.
A second, independently drawn set of $64$ graphs serves as validation, and we select the checkpoint by validation AUC.

\begin{table*}[b]
\centering
\footnotesize
\setlength{\tabcolsep}{3pt}
\caption{\textbf{The evaluation families fix one canonical setting per topology, the curated suite scatters structure across size, density, and each family's own parameter, and the real networks span the same range.} Summary statistics for every network: $N$ nodes, $|E|$ edges, $\langle k\rangle$ mean degree, $k_{\min}$ and $k_{\max}$ degree extremes, $\mathrm{CV}_k$ degree coefficient of variation (standard deviation over mean), $C$ average local clustering, and $\langle\ell\rangle$ average shortest-path length. Synthetic evaluation rows average five realizations per family; curated rows report the range over the $16$ graphs in each family. Clustering and $\langle\ell\rangle$ are estimated from random node samples, and real-world statistics use the largest connected component.}
\label{tab:networks}
\begin{tabular}{llrrrrrrrr}
\toprule
Network & Param. & $N$ & $|E|$ & $\langle k\rangle$ & $k_{\min}$ & $k_{\max}$ & $\mathrm{CV}_k$ & $C$ & $\langle\ell\rangle$ \\
\midrule
\multicolumn{10}{l}{\textit{Synthetic evaluation families ($N=5000$, one fixed setting each)}}\\
\textsc{ER}       & $-$                       & $5{,}000$ & $29{,}900$ & $12$ & $2$ & $27$  & $0.29$ & $0.00$ & $3.7$  \\
\textsc{PowL}     & $\gamma{=}2.5$            & $5{,}000$ & $30{,}000$ & $12$ & $4$ & $305$ & $1.50$ & $0.02$ & $3.2$  \\
\textsc{ClusPowL} & $\gamma{=}2.5,\,C{\approx}0.15$ & $5{,}000$ & $29{,}600$ & $12$ & $3$ & $280$ & $1.44$ & $0.15$ & $3.3$  \\
\textsc{CorPer}   & core frac $0.2$           & $5{,}000$ & $30{,}000$ & $12$ & $1$ & $119$ & $0.90$ & $0.01$ & $3.5$  \\
\textsc{RandGeo}  & radius ${\approx}0.028$   & $5{,}000$ & $30{,}100$ & $12$ & $1$ & $25$  & $0.30$ & $0.60$ & $24.0$ \\
\midrule
\multicolumn{10}{l}{\textit{Real-world networks (largest connected component)}}\\
US power grid & $-$ & $4{,}941$  & $6{,}594$  & $2.67$  & $1$ & $19$    & $0.67$ & $0.08$ & $19.0$ \\
Oregon AS     & $-$ & $10{,}670$ & $22{,}002$ & $4.12$  & $1$ & $2{,}312$ & $7.76$ & $0.30$ & $3.6$  \\
Chicago road  & $-$ & $12{,}979$ & $20{,}627$ & $3.18$  & $1$ & $7$     & $0.35$ & $0.04$ & $42.6$ \\
OpenFlights   & $-$ & $3{,}188$  & $18{,}833$ & $11.81$ & $1$ & $248$   & $2.13$ & $0.49$ & $3.9$  \\
AS1221        & $-$ & $3{,}515$  & $4{,}322$  & $2.46$  & $1$ & $103$   & $2.24$ & $0.02$ & $6.4$  \\
\midrule
\multicolumn{10}{l}{\textit{Curated training suite (ranges over $16$ graphs per family; \textsc{ClusPowL} excluded)}}\\
\textsc{ER}      & $-$                        & $2{,}582$--$14{,}714$ & $4{,}909$--$56{,}604$   & $2.8$--$12.5$ & $1$--$2$ & $10$--$26$    & $0.28$--$0.52$ & $0.00$          & $3.5$--$9.7$   \\
\textsc{PowL}    & $\gamma\,2.1$--$3.2$       & $2{,}538$--$14{,}378$ & $4{,}637$--$76{,}337$   & $3.0$--$12.4$ & $1$--$2$ & $71$--$2{,}046$ & $1.10$--$4.34$ & $0.00$--$0.20$ & $2.9$--$5.9$   \\
\textsc{CorPer}  & core frac $0.11$--$0.45$   & $2{,}544$--$13{,}278$ & $4{,}742$--$100{,}205$  & $3.2$--$15.6$ & $1$      & $15$--$209$   & $0.67$--$1.72$ & $0.00$--$0.08$ & $3.1$--$7.7$   \\
\textsc{RandGeo} & radius $0.011$--$0.034$    & $2{,}008$--$12{,}606$ & $5{,}376$--$67{,}823$   & $4.6$--$10.8$ & $1$      & $13$--$25$    & $0.31$--$0.43$ & $0.56$--$0.60$ & $21.3$--$162.8$ \\
\bottomrule
\end{tabular}
\end{table*}

\paragraph{Load distributions.}
The \emph{uniform} loads are drawn from a bounded uniform distribution around the target mean, giving moderate variation without a heavy tail or distinct groups.
The \emph{Pareto} loads are drawn from a heavy-tailed distribution scaled so its expectation equals the target mean, so a small number of nodes carry much larger loads than the rest.
The \emph{bimodal} loads are drawn from a two-group mixture with a smaller high-load group and a larger low-load group, a structured heterogeneity distinct from the gradual Pareto tail.

\subsection{Node Features}
\label{app:features}

The policy sees each node through a six-dimensional feature vector $\mathbf{x}_v$ that summarizes the node and its one-hop neighborhood, the locality the cascade redistributes over.
For a node $v$ with load $\ell_v$, degree $d_v=|\mathcal{N}(v)|$, and neighborhood $\mathcal{N}(v)$, the entries are the log-transformed own load $\log(\ell_v{+}\epsilon)$, degree $\log(d_v{+}1)$, mean neighbor load $\log(\bar\ell_v{+}\epsilon)$, maximum neighbor load $\log(\ell^{\max}_v{+}\epsilon)$, neighbor-load variance $\log(\mathrm{Var}_v{+}1)$, and two-hop degree $\log(d^{(2)}_v{+}1)$, where
\[
\bar\ell_v=\tfrac1{d_v}\!\sum_{u\in\mathcal{N}(v)}\!\ell_u,\quad
\ell^{\max}_v=\!\max_{u\in\mathcal{N}(v)}\!\ell_u,\quad
\mathrm{Var}_v=\tfrac1{d_v}\!\sum_{u\in\mathcal{N}(v)}\!(\ell_u-\bar\ell_v)^2,
\]
and $d^{(2)}_v=\sum_{u\in\mathcal{N}(v)}d_u$ is the summed degree of $v$'s neighbors.
The log transform compresses the heavy-tailed loads and degrees into a comparable range, with $\epsilon=10^{-8}$ keeping it finite at zero; isolated nodes take the four neighborhood features as zero before the log.
The mean and maximum neighbor loads capture the two ways a node is stressed under local redistribution, a steady inflow from many neighbors or a spike from one failing hub, and the two-hop degree signals how far a local failure can spread in a second round.

\subsection{GNN Architecture}
\label{app:gnn}

The policy is a message-passing GNN in the GraphSAGE family~\cite{hamilton2017inductive}, with degree-normalized mean aggregation~\cite{kipf2017gcn} and a residual update at each layer.
It first maps each feature vector to a hidden state,
\begin{equation}
  \mathbf{h}_v^{(0)} = \psi_{\mathrm{in}}(\mathbf{x}_v),
\end{equation}
with $\psi_{\mathrm{in}}$ a linear map into width $h$.
At layer $i$, each node collects messages from its neighbors, normalized by the sender's degree so that it parallels the load redistribution logic in the cascade,
\begin{align}
  \mathbf{m}_{u\to v}^{(i)} &= \frac{\mathbf{h}_u^{(i)}}{\max\{d_u,1\}},
  \\
  \mathbf{a}_v^{(i)} &= \sum_{u\in\mathcal{N}(v)} \mathbf{m}_{u\to v}^{(i)},
\end{align}
and updates its state with an MLP on the concatenation of the current state and the aggregate, wrapped in a residual connection,
\begin{equation}
  \mathbf{h}_v^{(i+1)} = \mathrm{MLP}_{i}\!\left(\mathbf{h}_v^{(i)} \,\|\, \mathbf{a}_v^{(i)}\right) + \mathbf{h}_v^{(i)},
  \label{eq:gnn_update}
\end{equation}
where $\|$ is concatenation.
The residual connection lets the network add depth without the oversmoothing that would otherwise pull neighboring node states together as layers stack.
After $K$ layers a linear head reads out one scalar residual logit per node,
\begin{equation}
  \mu_{\theta,v} = \psi_{\mathrm{out}}\!\left(\mathbf{h}_v^{(K)}\right),
\end{equation}
collected into $\boldsymbol{\mu}_{\theta}(G,\mathbf{X})=(\mu_{\theta,v})_{v\in\mathcal{V}}$, where $\theta$ gathers all layer parameters.
The head $\psi_{\mathrm{out}}$ is zero-initialized, so every residual logit starts at zero and the policy begins training exactly at the anchor allocation of Section~\ref{subsec:residual_softmax}.

\subsection{Reward Estimation and Training}
\label{app:training}

\paragraph{Reward grid.}
The failure-fraction grid $\mathcal{P}$ is chosen during training to span the range over which the anchor allocation's survival curve falls from intact to collapsed; failure fractions outside this band separate allocations only weakly and add variance without signal.
For each $p\in\mathcal{P}$ the $M$ initial failure sets $\mathcal{F}_{p,1},\dots,\mathcal{F}_{p,M}$ are drawn uniformly among node sets of size $\lfloor pN\rfloor$ and pre-sampled once per training grid, so that every sampled allocation in a batch is compared against the same failure sets and reward differences reflect the allocation rather than the draw.

\begin{table*}[b]
\centering
\caption{Transfer as a source-to-target matrix (absolute AUC, uniform load, $B{=}0.75$). Among synthetics, the diagonal is within-family transfer (train on eight realizations, test on two held out); off-diagonal cells are cross-family. \methodpre is pre-trained on a curated suite of $64$ synthetic graphs and applied to every target without further training; \method is trained on the target itself. Both are shown for each training anchor (\textsc{Uniform} and \textsc{LR-FSA}); the main text reports the better of the two. The bottom row is the best heuristic on each target. Boldface marks the column maximum among learned policies.}
\label{tab:transfer_uniform_auc}
\setlength{\tabcolsep}{3.5pt}
\begin{tabular}{lrrrrr|rrrrr}
\toprule
 & \multicolumn{5}{c|}{Synthetic families} & \multicolumn{5}{c}{Real-world networks} \\
\cmidrule(lr){2-6}\cmidrule(l){7-11}
Source & \textsc{ER} & \textsc{PowL} & \textsc{ClusPowL} & \textsc{CorPer} & \textsc{RandGeo}
       & US grid & Oregon AS & Chicago & OpenFl. & AS1221 \\
\midrule
\textsc{ER}         & 0.373 & 0.309 & 0.288 & 0.324 & 0.146
                    & 0.033 & 0.010 & 0.023 & 0.069 & 0.046 \\
\textsc{PowL}       & 0.374 & 0.356 & 0.329 & 0.344 & 0.154
                    & 0.030 & 0.056 & 0.021 & 0.199 & 0.166 \\
\textsc{ClusPowL}   & 0.373 & 0.353 & 0.329 & 0.342 & 0.156
                    & 0.027 & 0.033 & 0.018 & 0.215 & 0.141 \\
\textsc{CorPer}     & 0.374 & 0.346 & 0.322 & 0.348 & 0.155
                    & 0.029 & 0.010 & 0.022 & 0.183 & 0.090 \\
\textsc{RandGeo}    & 0.371 & 0.241 & 0.229 & 0.280 & \textbf{0.168}
                    & 0.030 & 0.038 & \textbf{0.024} & 0.030 & 0.038 \\
\midrule
\methodpre (Unif.\ anchor) & 0.370 & 0.341 & 0.313 & 0.335 & 0.143 & 0.040 & \textbf{0.336} & 0.016 & 0.238 & 0.239 \\
\methodpre (\textsc{LR-FSA} anchor) & \textbf{0.375} & \textbf{0.357} & \textbf{0.331} & \textbf{0.348} & 0.163 & \textbf{0.042} & 0.224 & 0.023 & 0.261 & 0.209 \\
\midrule
\method (Unif.\ anchor) & 0.367 & 0.329 & 0.317 & 0.333 & 0.154 & 0.033 & 0.319 & 0.012 & \textbf{0.262} & \textbf{0.286} \\
\method (\textsc{LR-FSA} anchor) & 0.374 & 0.355 & 0.330 & 0.347 & 0.158 & 0.034 & 0.074 & 0.020 & 0.251 & 0.230 \\
\midrule
Best heuristic      & 0.362 & 0.303 & 0.273 & 0.297 & 0.132
                    & 0.022 & 0.237 & 0.019 & 0.076 & 0.263 \\
\bottomrule
\end{tabular}
\end{table*}

\paragraph{Policy gradient.}
We optimize the objective~\eqref{eq:policy_objective} with REINFOR-CE~\cite{williams1992}.
The score-function identity gives
\begin{equation}
  \nabla_{\theta}\mathcal{J}(\theta)
  =
  \mathbb{E}_{\mathbf{z}\sim p_{\theta}}
  \left[
  \widehat{\mathrm{AUC}}
  \left(
  G,\mathbf{l},\mathbf{s}(\mathbf{z})
  \right)
  \nabla_{\theta}\log p_{\theta}(\mathbf{z}\mid G,\mathbf{X})
  \right],
  \label{eq:reinforce_gradient}
\end{equation}
so the simulator enters only through the scalar reward, and the gradient flows through the log-probability of the sampled residual logits, which is differentiable in $\theta$.
At each iteration we sample $Q$ residual-logit vectors $\mathbf{z}^{(1)},\dots,\mathbf{z}^{(Q)}$, each yielding an allocation $\mathbf{s}^{(b)}$ and reward $R^{(b)}=\widehat{\mathrm{AUC}}(G,\mathbf{l},\mathbf{s}^{(b)})$, and subtract the batch mean as a baseline,
\begin{equation}
  A^{(b)} = R^{(b)}-\bar{R},\qquad \bar{R}=\frac{1}{Q}\sum_{b=1}^{Q}R^{(b)}.
\end{equation}
The training loss combines the baselined policy-gradient term with a small L2 penalty on the residual logits,
\begin{equation}
  \mathcal{L}_{\mathrm{RL}}(\theta)
  =
  -
  \frac{1}{Q}
  \sum_{b=1}^{Q}
  A^{(b)}
  \log p_{\theta}\left(\mathbf{z}^{(b)}\mid G,\mathbf{X}\right)
  +
  \frac{\lambda}{N}
  \left\|
  \boldsymbol{\mu}_{\theta}(G,\mathbf{X})
  \right\|_2^2 ,
  \label{eq:reinforce_loss}
\end{equation}
where the penalty keeps the residual logits from drifting far from the anchor and stabilizes the policy-gradient updates.

\paragraph{Hyperparameters.}
The policy uses $K=3$ message-passing layers of hidden width $64$, each update a two-layer ReLU MLP with a residual connection, and a zero-initialized linear readout.
We train each instance with Adam~\cite{kingma2015adam} at learning rate $10^{-3}$ for $200$ iterations, drawing $Q=32$ sampled allocations per iteration and annealing the exploration scale $\sigma$ linearly from $0.3$ to $0.1$, with an L2 penalty $\lambda=10^{-4}$ on the residual logits.
Training rewards use $32$ failure samples per failure fraction on a $41$-point grid covering the informative band of the anchor's survival curve; reported AUC uses $100$ samples on a fixed $101$-point grid over $p\in[0,0.5]$.
We draw the evaluation failure sets independently of the training sets and fix them across all methods, so reported AUC scores every allocation on the same unseen failures.
We replicate training over five seeds per configuration (realization, load, budget, anchor) and report AUC averaged over the five; the per-cell standard deviation in \method AUC is $\approx0.001$--$0.010$.

\paragraph{Use of generative AI}
We used generative AI assistants (Cursor and Claude) for editing prose, refactoring code, and preparing reproducibility packaging.
All scientific claims, experimental design, results, and final wording were reviewed and verified by the authors.

\section{Additional Results and Analyses}
\label{app:results}
\begin{figure}[h]
    \centering
    \includegraphics[width=0.6\columnwidth]{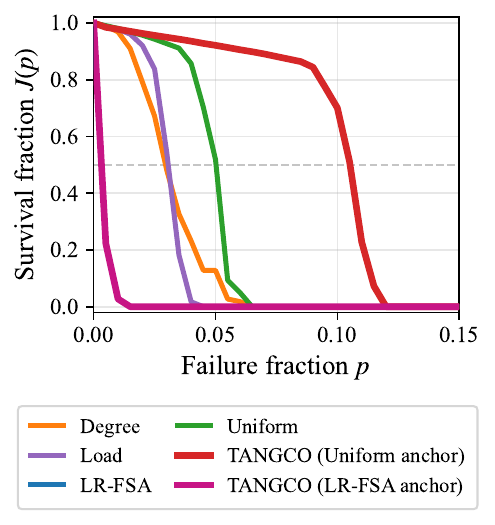}
    \caption{\textbf{Keeping both anchors recovers the strongest allocation when one collapses.} Survival fraction $J(p)$ vs failure fraction $p$ on Oregon AS under Pareto load at $B=0.5$: the \textsc{LR-FSA} heuristic collapses while \textsc{Uniform} holds, so the \textsc{LR-FSA}-anchored policy collapses with it, while the \textsc{Uniform}-anchored policy recovers the strongest allocation. Keeping the better of the two anchors yields the reported \method.}
    \label{fig:oregon_anchor}
\end{figure}
\subsection{Anchor Selection}
\label{app:anchor}
Figure~\ref{fig:oregon_anchor} illustrates the anchor-failure mode that motivates keeping both anchors: on Oregon AS under Pareto load at $B=0.5$, the \textsc{LR-FSA} heuristic collapses while \textsc{Uniform} holds, and optimizing over both anchors recovers the strongest allocation.
Across the full suite, \textsc{LR-FSA} anchors the winning policy on every synthetic family under uniform and bimodal loads (typically $10/10$ realizations).

The two anchors are closest under Pareto load, where \textsc{Uniform} becomes competitive at tight budgets, winning entirely on \textsc{CorPer} ($10/10$ at $B=0.5$) and on a majority of \textsc{ER} realizations.
On the real networks the \textsc{Uniform} anchor supplies the winning policy on both hub-heavy AS topologies, Oregon and AS1221, in all nine conditions each, while \textsc{LR-FSA} anchors every winning policy on the near-planar Chicago network.
The best hand-designed baseline also shifts on the reals: \textsc{Degree} is the strongest heuristic on Rocketfuel AS1221 in all nine conditions, a rule that never wins on any synthetic family, yet \method still improves on it in all nine (by $+11\%$ on average).

\begin{table*}[t]
\centering
\caption{Same absolute-AUC transfer matrix as Table~\ref{tab:transfer_uniform_auc}, under Pareto load ($B{=}0.75$), with \methodpre and \method split by training anchor as in Table~\ref{tab:transfer_uniform_auc}. Boldface marks the column maximum among learned policies.}
\label{tab:transfer_pareto_auc}
\setlength{\tabcolsep}{3.5pt}
\begin{tabular}{lrrrrr|rrrrr}
\toprule
 & \multicolumn{5}{c|}{Synthetic families} & \multicolumn{5}{c}{Real-world networks} \\
\cmidrule(lr){2-6}\cmidrule(l){7-11}
Source & \textsc{ER} & \textsc{PowL} & \textsc{ClusPowL} & \textsc{CorPer} & \textsc{RandGeo}
       & US grid & Oregon AS & Chicago & OpenFl. & AS1221 \\
\midrule
\textsc{ER}         & 0.330 & 0.245 & 0.224 & 0.283 & 0.061
                    & 0.018 & 0.011 & 0.011 & 0.041 & 0.021 \\
\textsc{PowL}       & 0.330 & \textbf{0.319} & 0.292 & 0.312 & 0.095
                    & 0.020 & 0.053 & 0.011 & 0.216 & 0.156 \\
\textsc{ClusPowL}   & 0.330 & 0.318 & 0.293 & 0.313 & 0.092
                    & 0.020 & 0.022 & 0.011 & 0.194 & 0.126 \\
\textsc{CorPer}     & 0.330 & 0.310 & 0.285 & 0.312 & 0.075
                    & 0.021 & 0.010 & 0.011 & 0.122 & 0.021 \\
\textsc{RandGeo}    & 0.324 & 0.235 & 0.218 & 0.264 & 0.113
                    & 0.020 & 0.010 & 0.012 & 0.040 & 0.046 \\
\midrule
\methodpre (Unif.\ anchor) & 0.328 & 0.311 & 0.282 & 0.311 & 0.054 & 0.047 & 0.321 & 0.011 & 0.213 & 0.250 \\
\methodpre (\textsc{LR-FSA} anchor) & \textbf{0.332} & 0.317 & 0.291 & \textbf{0.314} & 0.065 & \textbf{0.060} & 0.120 & 0.011 & 0.210 & 0.219 \\
\midrule
\method (Unif.\ anchor) & 0.329 & 0.307 & 0.277 & 0.307 & 0.097 & 0.040 & \textbf{0.326} & 0.010 & \textbf{0.240} & \textbf{0.282} \\
\method (\textsc{LR-FSA} anchor) & 0.331 & 0.318 & \textbf{0.295} & 0.310 & \textbf{0.123} & 0.032 & 0.037 & \textbf{0.012} & 0.182 & 0.213 \\
\midrule
Best heuristic      & 0.320 & 0.269 & 0.238 & 0.262 & 0.083
                    & 0.020 & 0.243 & 0.012 & 0.082 & 0.260 \\
\bottomrule
\end{tabular}
\end{table*}

\subsection{Transferability}
\label{app:transfer}

This appendix details the two transfer settings of Section~\ref{sec:experiments:transfer} and reports the full per-target results.
In the per-graph setting, each family contributes ten independent realizations.
We train one policy jointly on eight of them for $400$ iterations and evaluate it on the two held out, which forms the within-family diagonal of Table~\ref{tab:transfer}.
For the cross-family entries, we take each per-family policy and apply it, without retraining, to the held-out graphs of every other family.
Tables~\ref{tab:transfer_uniform_auc} and~\ref{tab:transfer_pareto_auc} report every policy, including \methodpre, as absolute AUC under both loads.

\methodpre is pre-trained on the curated suite of $64$ synthetic graphs described in Appendix~\ref{app:graphs}, and sees no real network or test graph during training.
The test targets are never seen in training: the held-out realizations of the five evaluation families and the five real networks.
Because one policy must fit many graphs at once, pre-training runs longer than per-instance training, $6{,}400$ iterations against $200$; all other settings match Appendix~\ref{app:training}.

Under Pareto load the transfer picture matches uniform load.
Within-family transfer stays positive on all five families and lands within $0.001$ AUC of per-instance training on four of them.
Cross-family transfer improves on the target heuristic in $14$ of $20$ cells, against $17$ under uniform load.
The six failures are the pairs whose structure differs most: a random-geometric policy applied to the heavy-tailed families, an \textsc{ER} policy applied to the heavy-tailed and spatial families, and a core-periphery policy applied to \textsc{RandGeo}.
\textsc{RandGeo} is the one family whose diagonal moves between loads: a shared within-family policy exceeds per-instance training by $0.010$ AUC under uniform load, while per-instance training is ahead by the same margin under Pareto load.

\begin{figure}[b]
\centering
\includegraphics[width=0.8\linewidth]{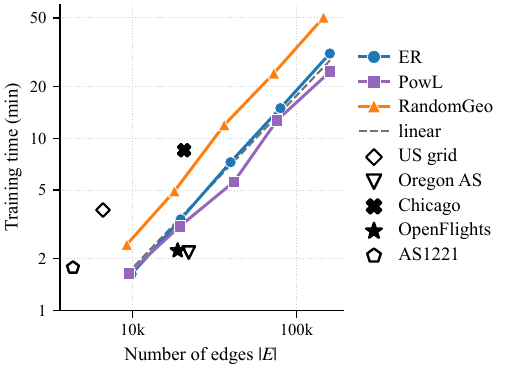}
\caption{\textbf{Sparse networks with long cascades take longer than edge count predicts.} Median training time versus number of edges $|E|$ (log--log, $200$ iterations), for three synthetic families up to $40{,}000$ nodes and the five evaluated real networks. Fixed mean degree makes $|E|\propto N$ for the synthetic families, so this view rescales Figure~\ref{fig:teaser}(c); US grid and AS1221 sit above the trend because their deeper cascades dominate the $O(N\bar{t})$ cost term.}
\label{fig:scaling_edges}
\end{figure}

\begin{figure*}[t]
\centering
\includegraphics[width=0.8\textwidth]{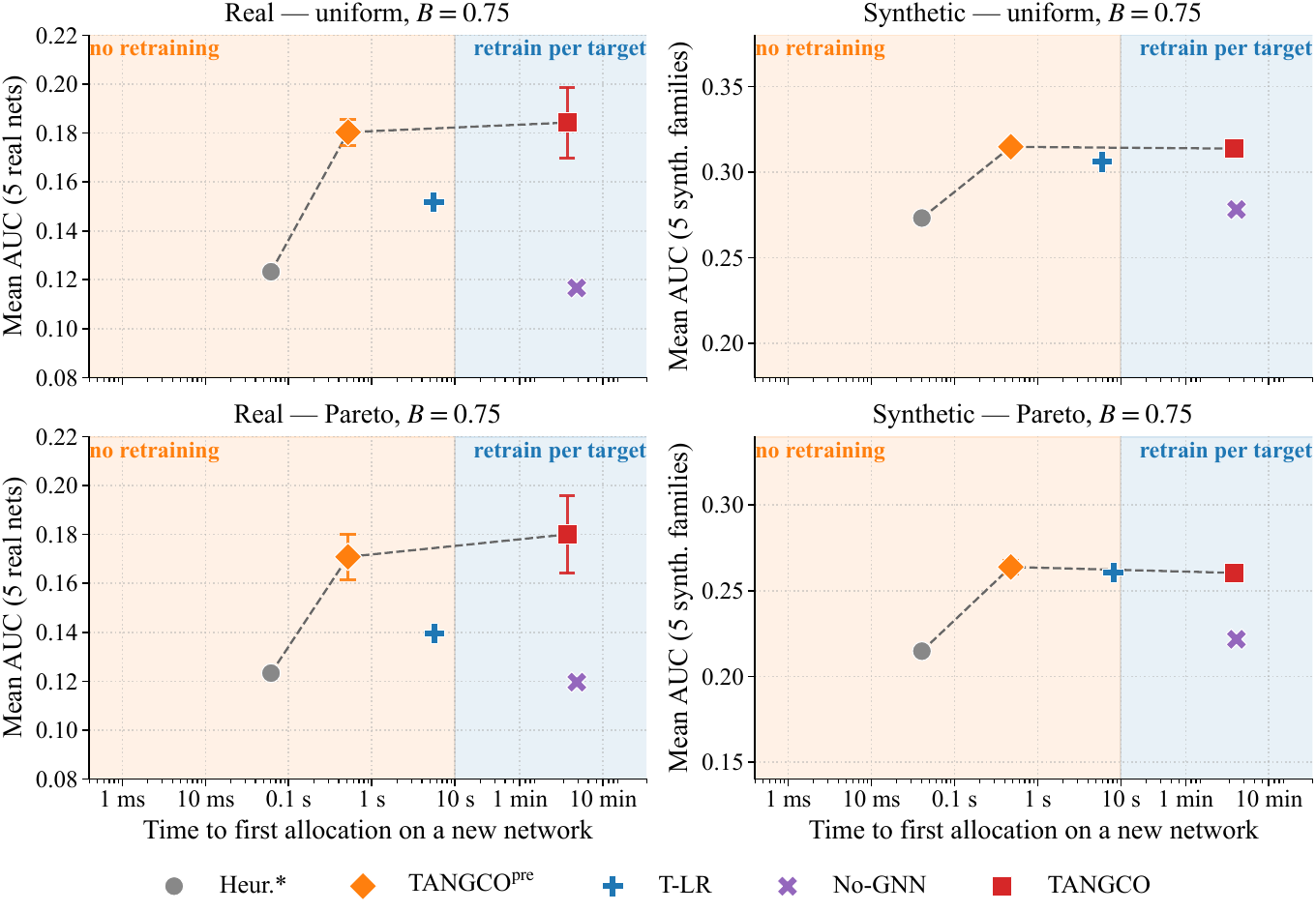}
\caption{\textbf{\methodpre reaches near-instance robustness at heuristic deployment cost under both load distributions.} Mean AUC versus time to first allocation on a new network (log $x$-axis), for the five real networks and five synthetic families under uniform and Pareto load at $B=0.75$. Figure~\ref{fig:teaser}(b) shows the uniform-load front. Left band: methods that need no per-target training (best heuristic, \tlr, \methodpre); right band: methods retrained per target (\nognn, instance \method). Error bars on \methodpre and instance \method are the mean across networks of the within-network seed standard deviation ($K{=}5$); on synthetic panels those values are $\lesssim 0.005$ AUC and fall below visual resolution at the plotted $y$-scale.}
\label{fig:pareto_full}
\end{figure*}

\subsection{Scalability Details}
\label{app:scaling}

The policy adds one forward and one backward pass per training iteration, each $O\!\left(K(Eh + Nh^2)\right)$ for $K$ message-passing layers of width $h$: message passing contributes $O(Eh)$ and the per-node update $O(Nh^2)$ per layer.
At the settings of Section~\ref{sec:experiments:setup} ($K=3$, $h=64$) this is negligible beside the $\approx8.4$M cascade simulations per run, each $O(N\bar{t}+E)$, as Figure~\ref{fig:teaser}(c) confirms.

At every network size the cascade simulator accounts for over $94\%$ of the training time and the GNN forward and backward passes for under $2\%$. A full $200$-iteration run takes $24$ to $50$ minutes at $40{,}000$ nodes on $16$ CPU cores, and under nine minutes on every evaluated network.

The cascade simulations are independent, and the implementation exploits this: the $M$ failure sets at a given allocation and failure fraction are simulated in parallel.
The $Q$ sampled allocations and the $|\mathcal{P}|$ failure fractions are likewise independent and are currently evaluated in sequence, so further parallelism is available without changing the method.

Figure~\ref{fig:pareto_full} extends the deployment comparison to both load distributions. Figure~\ref{fig:scaling_edges} redraws the size-scaling result against edge count. The synthetic families hold mean degree fixed, so their edge count is proportional to node count and the three curves keep their near-linear slopes. The real networks separate by cascade depth. Sparse networks such as US grid and AS1221 sit above the trend: their longer cascades inflate the $O(N\bar{t})$ term that dominates each simulation (Section~\ref{subsec:cascade_dynamics}). Edge count alone underpredicts their cost, which confirms that the node-sweep term drives runtime.

\subsection{No-GNN Ablation}
\label{app:ablation}
The \nognn variant tests whether reward-driven search alone can match \method without message passing.
It replaces the GNN by a free residual logit $\Delta\mu_v$ per node, zero-initialized so that the deterministic allocation matches the training anchor at the start of training, and feeds that residual into the same residual-softmax allocation head as \method (Section~\ref{subsec:residual_softmax}).
Training uses the same REINFORCE loop, Adam schedule, exploration annealing, L2 penalty on residual logits, failure-sample counts, two anchors (\textsc{Uniform}, \textsc{LR-FSA}), five seeds, and $200$ iterations as the instance-\method runs.
The ablation covers all ten networks (five synthetic families and five reals); one realization represents each family, and both learned policies are five-seed means.
We report the better of the two anchors, taking the best-validation-AUC checkpoint per seed as in the primary protocol (Section~\ref{subsec:reinforce_training}).
Table~\ref{tab:ablation_full} collects the uniform and Pareto results at $B=0.75$; the body Table~\ref{tab:ablation} shows the uniform block alone.
\method beats the \nognn variant in all $20$ cells (mean $+0.052$ AUC under uniform load, $+0.050$ under Pareto).
Message passing matters most on the hub-heavy real networks: on OpenFlights the \nognn variant barely clears the heuristic ($0.083$ vs.\ $0.076$) while \method more than triples it ($0.262$).
On AS1221 the \textsc{Degree} heuristic ($0.263$) outranks both search variants, which improve only modestly on the shared \textsc{Uniform} anchor ($0.213$ and $0.235$ from $0.205$); \method moves further to $0.286$ and alone surpasses it.
We test whether this result is specific to the REINFORCE optimizer by optimizing the same residual-softmax head with CMA-ES~\cite{hansen2016cma}, a derivative-free evolutionary search, using a free per-node logit vector zero-initialized at the anchor, two anchors, five seeds, and the same AUC reward.
Full CMA-ES maintains an $N\times N$ covariance that is infeasible for networks of thousands of nodes, so we use the separable (diagonal) variant, whose covariance scales linearly in $N$~\cite{ros2008sepcma}.
We cap the search at $6{,}400$ reward evaluations, matching the \nognn REINFORCE budget ($200$ iterations $\times$ $32$ samples), with initial step size $\sigma_0=0.2$, and keep the best-validation-AUC allocation as in the primary protocol.
Table~\ref{tab:ablation_full} shows CMA-ES tracks the \nognn variant closely on most cells (within $0.015$ AUC on $15$ of $20$) and stays near the best heuristic (mean $\Delta$AUC $+0.009$ under uniform load, $+0.012$ under Pareto).
\method beats it in all $20$ cells (mean $+0.042$ and $+0.039$).
Reward-driven search, whether policy-gradient or evolutionary, cannot recover the gain that message passing provides.

\begin{table}[t]
\centering
\caption{\textbf{\method beats the \nognn variant in all $20$ family-load cells; \nognn alone stays close to the best heuristic.} AUC at $B=0.75$, single realization per family, uniform and Pareto loads. \emph{Heuristic*} is the best of the four heuristics. The learned search variants (\nognn and CMA-ES, a derivative-free optimizer) are five-seed means (best of two anchors); \method values are bold.}
\label{tab:ablation_full}
\small
\setlength{\tabcolsep}{4pt}
\begin{tabular}{llrrrr}
\toprule
Load & Graph & \method & \nognn & CMA-ES & Heuristic* \\
\midrule
Uniform & \textsc{ER}       & $\mathbf{0.372}$ & $0.358$ & $0.357$ & $0.357$ \\
        & \textsc{PowL}     & $\mathbf{0.354}$ & $0.311$ & $0.313$ & $0.302$ \\
        & \textsc{ClusPowL} & $\mathbf{0.327}$ & $0.279$ & $0.282$ & $0.270$ \\
        & \textsc{CorPer}   & $\mathbf{0.345}$ & $0.299$ & $0.295$ & $0.295$ \\
        & \textsc{RandGeo}  & $\mathbf{0.171}$ & $0.144$ & $0.141$ & $0.141$ \\
        & US grid           & $\mathbf{0.034}$ & $0.024$ & $0.022$ & $0.022$ \\
        & Oregon AS         & $\mathbf{0.319}$ & $0.244$ & $0.259$ & $0.237$ \\
        & Chicago           & $\mathbf{0.020}$ & $0.019$ & $0.019$ & $0.019$ \\
        & OpenFlights       & $\mathbf{0.262}$ & $0.083$ & $0.145$ & $0.076$ \\
        & AS1221            & $\mathbf{0.286}$ & $0.213$ & $0.235$ & $0.263$ \\
\midrule
Pareto  & \textsc{ER}       & $\mathbf{0.314}$ & $0.286$ & $0.283$ & $0.282$ \\
        & \textsc{PowL}     & $\mathbf{0.302}$ & $0.259$ & $0.265$ & $0.249$ \\
        & \textsc{ClusPowL} & $\mathbf{0.275}$ & $0.229$ & $0.237$ & $0.220$ \\
        & \textsc{CorPer}   & $\mathbf{0.303}$ & $0.251$ & $0.252$ & $0.247$ \\
        & \textsc{RandGeo}  & $\mathbf{0.109}$ & $0.083$ & $0.076$ & $0.075$ \\
        & US grid           & $\mathbf{0.040}$ & $0.022$ & $0.027$ & $0.020$ \\
        & Oregon AS         & $\mathbf{0.326}$ & $0.251$ & $0.266$ & $0.243$ \\
        & Chicago           & $\mathbf{0.012}$ & $0.012$ & $0.012$ & $0.012$ \\
        & OpenFlights       & $\mathbf{0.240}$ & $0.089$ & $0.151$ & $0.082$ \\
        & AS1221            & $\mathbf{0.282}$ & $0.225$ & $0.244$ & $0.260$ \\
\bottomrule
\end{tabular}
\end{table}

\subsection{Message-Passing Depth}
\label{app:depth_anchor}
The depth sweep varies the number of message-passing layers over $K\in\{0,\dots,5\}$ on the five synthetic families plus the US grid and Oregon AS, under the same one-realization / five-seed protocol as Appendix~\ref{app:ablation}.
Table~\ref{tab:depth_anchor} reports each anchored policy's relative improvement over its \emph{own} anchor, which isolates what message passing must learn from what the anchor already provides.
Under the \textsc{Uniform} anchor, which carries no structural signal, the $K{=}0\!\to\!K{=}1$ step is large: one message-passing hop supplies the neighborhood correction, after which $K=2$--$5$ stay flat on the synthetic families.
Under the \textsc{LR-FSA} anchor the same step nearly vanishes, because the anchor already injects the one-hop local-risk estimate.
We omit Oregon AS under \textsc{LR-FSA}: the prior collapses on its extreme hubs, leaving a near-zero denominator that makes the relative figure uninterpretable.
The US grid and Oregon AS keep fluctuating at higher $K$, though their near-floor AUC inflates the relative figures.

\begin{table}[b]
\centering
\caption{\textbf{The $K{=}0\!\to\!K{=}1$ step is large under the \textsc{Uniform} anchor and small under the \textsc{LR-FSA} anchor, which already carries the one-hop signal.} Depth split by training anchor (uniform load, $B=0.75$): relative improvement in AUC of each anchored policy over its own anchor (\%).}
\label{tab:depth_anchor}
\begin{tabular}{lrrrrrr}
\toprule
Network & $K{=}0$ & $K{=}1$ & $K{=}2$ & $K{=}3$ & $K{=}4$ & $K{=}5$ \\
\midrule
\multicolumn{7}{l}{\emph{\textsc{Uniform} anchor}} \\
\textsc{ER}       & $+3.9$  & $+6.8$  & $+7.3$  & $+7.0$  & $+7.1$  & $+7.3$ \\
\textsc{PowL}     & $+35.9$ & $+62.2$ & $+61.3$ & $+60.7$ & $+61.3$ & $+60.0$ \\
\textsc{ClusPowL} & $+29.8$ & $+56.3$ & $+56.7$ & $+56.8$ & $+56.3$ & $+56.5$ \\
\textsc{CorPer}   & $+39.5$ & $+57.1$ & $+57.6$ & $+58.0$ & $+57.4$ & $+57.7$ \\
\textsc{RandGeo}  & $+6.5$  & $+13.8$ & $+13.2$ & $+14.0$ & $+14.6$ & $+13.5$ \\
US grid           & $+63.0$ & $+88.2$ & $+74.2$ & $+89.1$ & $+77.0$ & $+92.9$ \\
Oregon AS         & $+24.1$ & $+39.4$ & $+32.6$ & $+34.4$ & $+35.0$ & $+23.9$ \\
\midrule
\multicolumn{7}{l}{\emph{\textsc{LR-FSA} anchor}} \\
\textsc{ER}       & $+3.3$  & $+3.7$  & $+3.8$  & $+4.0$  & $+3.9$  & $+3.8$ \\
\textsc{PowL}     & $+14.6$ & $+17.3$ & $+17.0$ & $+17.1$ & $+16.8$ & $+16.8$ \\
\textsc{ClusPowL} & $+18.2$ & $+21.0$ & $+21.2$ & $+21.0$ & $+20.9$ & $+20.9$ \\
\textsc{CorPer}   & $+15.4$ & $+17.4$ & $+17.1$ & $+17.0$ & $+17.3$ & $+17.3$ \\
\textsc{RandGeo}  & $+31.8$ & $+33.9$ & $+33.6$ & $+34.4$ & $+32.2$ & $+33.3$ \\
US grid           & $+30.1$ & $+61.4$ & $+74.3$ & $+58.6$ & $+50.6$ & $+66.8$ \\
Oregon AS         & \multicolumn{6}{c}{\emph{collapsed prior, omitted}} \\
\bottomrule
\end{tabular}
\end{table}

\newpage
\subsection{Per-Cell Fits for the Regime Map}
\label{app:regime}

Table~\ref{tab:regime_full} gives the nested fits behind the three regimes of Section~\ref{subsec:interp_signal}. Each column adds one local feature to the regression of $\log\!\left(s_v/(B/N)\right)$ on $\log r_v$, and a cell is \emph{linear} once some model on the ladder reaches $R^2\ge0.90$: regime A if risk alone does, regime B if the full linear model does, and regime C otherwise. On the regime-C cells a nonlinear additive fit in risk and degree recovers far more than any linear model: $0.98$ and $0.91$ on Oregon AS under uniform and Pareto load, $0.91$ and $0.88$ on AS1221, and $0.82$ on the US grid under Pareto load, against linear fits of $0.03$--$0.33$.

Cells are absent where the policy stays at its anchor and leaves no learned allocation to explain, which at $B=0.5$ is the feasibility floor of Section~\ref{sec:experiments:regime}.

\begin{table}[b]
\centering
\small
\caption{\textbf{Local-risk alone explains the allocation on ten of the nineteen cells; degree closes most of the remaining gap.} Nested $R^2$ for the regime map (uniform and Pareto loads, $B=0.75$). Columns add features cumulatively to $\log r$. \emph{Needs} names the feature closing most of the gap for regime-B cells.}
\label{tab:regime_full}
\begin{tabular}{llrrrrcl}
\toprule
Network & Load & $r$ & $+d$ & $+\ell$ & $+k$ & Reg. & Needs \\
\midrule
\textsc{ER}          & Uniform  & $0.95$ & $0.97$ & $0.99$ & $0.99$ & A & -- \\
                     & Pareto   & $0.78$ & $0.82$ & $0.93$ & $0.94$ & B & load \\
\textsc{PowL}        & Uniform  & $0.98$ & $0.98$ & $0.98$ & $0.98$ & A & -- \\
                     & Pareto   & $0.93$ & $0.96$ & $0.98$ & $0.98$ & A & -- \\
\textsc{ClusPowL}    & Uniform  & $0.97$ & $0.98$ & $0.98$ & $0.98$ & A & -- \\
                     & Pareto   & $0.95$ & $0.97$ & $0.98$ & $0.98$ & A & -- \\
\textsc{CorPer}      & Uniform  & $0.98$ & $0.99$ & $0.99$ & $0.99$ & A & -- \\
                     & Pareto   & $0.91$ & $0.94$ & $0.95$ & $0.97$ & A & -- \\
\textsc{RandGeo}     & Uniform  & $0.40$ & $0.95$ & $0.96$ & $0.96$ & B & degree \\
                     & Pareto   & $0.83$ & $0.96$ & $0.97$ & $0.97$ & B & degree \\
\midrule
US grid              & Uniform  & $0.95$ & $0.96$ & $0.96$ & $0.97$ & A & -- \\
                     & Pareto   & $0.27$ & $0.27$ & $0.27$ & $0.49$ & C & -- \\
Oregon AS            & Uniform  & $0.00$ & $0.33$ & $0.33$ & $0.34$ & C & -- \\
                     & Pareto   & $0.01$ & $0.28$ & $0.28$ & $0.30$ & C & -- \\
Chicago              & Uniform  & $0.99$ & $0.99$ & $0.99$ & $1.00$ & A & -- \\
OpenFlights          & Uniform  & $0.96$ & $0.98$ & $0.98$ & $0.98$ & A & -- \\
                     & Pareto   & $0.83$ & $0.88$ & $0.88$ & $0.90$ & B & degree \\
AS1221               & Uniform  & $0.00$ & $0.03$ & $0.03$ & $0.06$ & C & -- \\
                     & Pareto   & $0.09$ & $0.19$ & $0.20$ & $0.21$ & C & -- \\
\bottomrule
\end{tabular}
\end{table}

Table~\ref{tab:regime_budget} repeats the classification at all three budgets. Eight of the nineteen cells change regime under a change of budget alone, holding the graph and the load fixed, which is why Section~\ref{subsec:interp_signal} attributes the regime to topology, load, and budget jointly rather than to topology.

\begin{figure*}[t]
    \centering
    \includegraphics[width=0.75\linewidth]{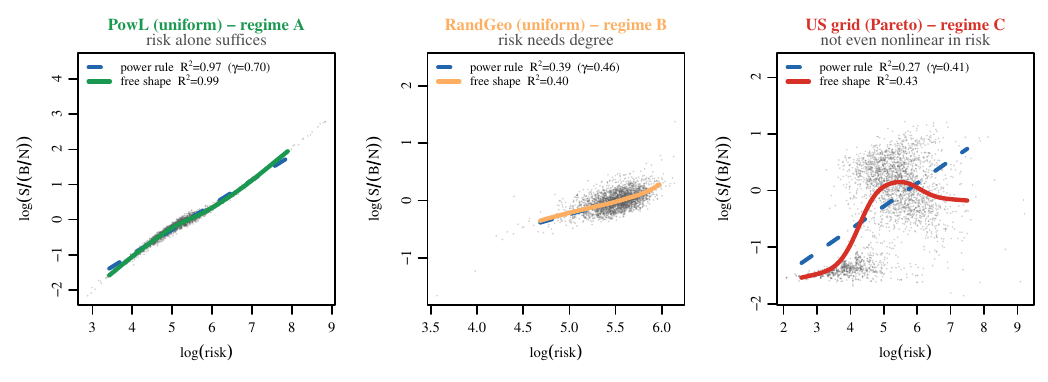}
    \caption{\textbf{Local-risk alone explains the learned allocation in regime A and fails in regimes B and C.} The learned allocation against local-risk, one network per regime (uniform load unless noted, $B=0.75$). Points are nodes; the dashed line is the power rule $s\propto r^{\gamma}$ (a straight line in these coordinates, slope $\gamma$), the solid curve is the same single feature with a free shape. We fit on every node and draw the curves over the range holding $99\%$ of them. \textbf{Left:} on \textsc{PowL} the power rule already fits, and freeing the shape adds nothing. \textbf{Middle:} on \textsc{RandGeo} neither fits, because risk is the wrong feature here and degree is what closes the gap. \textbf{Right:} on the US grid under Pareto load the power rule keeps raising capacity with risk while the policy levels off, and a free shape in risk alone still reaches only $0.43$.}
    \label{fig:gam_regimes}
\end{figure*}

Figure~\ref{fig:gam_regimes} shows one network per regime, fitting $\log s$ against $\log r$ alone, once as a power rule and once with a free shape. A single feature keeps the two fits comparable, since risk and degree are correlated on these graphs ($0.61$--$0.77$) and a model given both cannot attribute between them. The fitted power-rule exponent spans $0.57$ to $1.10$ across the regime-A networks, so no single value is right everywhere, which is why \tlr (Section~\ref{subsec:interp_rule}) fits a per-graph exponent rather than a shared one.
\tlr selects $\gamma$ on the training failure sets of Appendix~\ref{app:training} and reports AUC with $50$ samples per fraction on the same $101$-point evaluation grid (Table~\ref{tab:a3}); the learned policies use $100$ samples.

\begin{table}[b]
\centering
\small
\caption{\textbf{Eight of the nineteen cells change regime under a change of budget alone, holding graph and load fixed.} Regime at each budget; a check marks a changing cell. Dashes are cells where the policy stays at its anchor, leaving no learned allocation to explain.}
\label{tab:regime_budget}
\begin{tabular}{llcccc}
\toprule
Network & Load & $B{=}0.5$ & $B{=}0.75$ & $B{=}1.0$ & Changes \\
\midrule
\textsc{ER}          & Uniform  & B & A & A & $\checkmark$ \\
                     & Pareto   & B & B & A & $\checkmark$ \\
\textsc{PowL}        & Uniform  & A & A & A &  \\
                     & Pareto   & B & A & A & $\checkmark$ \\
\textsc{ClusPowL}    & Uniform  & A & A & A &  \\
                     & Pareto   & B & A & A & $\checkmark$ \\
\textsc{CorPer}      & Uniform  & A & A & A &  \\
                     & Pareto   & B & A & A & $\checkmark$ \\
\textsc{RandGeo}     & Uniform  & B & B & B &  \\
                     & Pareto   & C & B & B & $\checkmark$ \\
\midrule
US grid              & Uniform  & -- & A & C & $\checkmark$ \\
                     & Pareto   & -- & C & C &  \\
Oregon AS            & Uniform  & C & C & C &  \\
                     & Pareto   & C & C & C &  \\
Chicago              & Uniform  & -- & A & A &  \\
                     & Pareto   & -- & -- & A &  \\
OpenFlights          & Uniform  & A & A & A &  \\
                     & Pareto   & C & B & B & $\checkmark$ \\
AS1221               & Uniform  & C & C & C &  \\
                     & Pareto   & C & C & C &  \\
\bottomrule
\end{tabular}
\end{table}

\newpage
